\documentclass{article} % For LaTeX2e
\usepackage{iclr2023_conference,times}

\usepackage{hyperref}
\usepackage{url}

\usepackage{booktabs}
\usepackage{colortbl}
\usepackage{comment}

\usepackage{enumitem}
\usepackage{hyperref}
\usepackage[ruled,vlined]{algorithm2e}
\usepackage{algorithmic}

\usepackage{url}            % simple URL typesetting
\usepackage{amsfonts}       % blackboard math symbols
\usepackage{nicefrac}       % compact symbols for 1/2, etc.
\usepackage{microtype}      % microtypography
\usepackage{xcolor}         % colors
\usepackage{bbm}
\usepackage{amsmath}

\usepackage{dsfont} % indicator function

\usepackage{float,wrapfig}

\usepackage{array}
\usepackage{adjustbox}
\usepackage{multirow}
\usepackage{colortbl}
\usepackage{caption}
\usepackage{amsmath}
\usepackage[capitalise]{cleveref}
\usepackage{multicol}
\usepackage{mathtools}

\usepackage{soul}

\usepackage{xspace}
\usepackage[htt]{hyphenat}
\usepackage{changepage}

\makeatletter
\DeclareRobustCommand\onedot{\futurelet\@let@token\@onedot}
\def\@onedot{\ifx\@let@token.\else.\null\fi\xspace}

\def\eg{\emph{e.g}\onedot} 
\def\ie{\emph{i.e}\onedot}

\makeatother
\newcommand{\eqn}[1]{Equation~(\ref{#1})}
\newcommand{\fig}[1]{Figure~\ref{#1}}
\newcommand{\tbl}[1]{Table~\ref{#1}}

\newcommand{\ignore}[1]{}

\newcommand{\modelname}{PIC\xspace}

\definecolor{rowblue}{RGB}{220,230,240}
\definecolor{myorchid}{RGB}{150,10,30}
\definecolor{myblue}{RGB}{10,30,250}
\definecolor{mygreen}{RGB}{10,120,10}

\usepackage{amsmath,amsfonts,bm}

\def\eqref#1{equation~\ref{#1}}
\def\1{\bm{1}}

\DeclareMathAlphabet{\mathsfit}{\encodingdefault}{\sfdefault}{m}{sl}
\SetMathAlphabet{\mathsfit}{bold}{\encodingdefault}{\sfdefault}{bx}{n}

\DeclareMathOperator*{\argmin}{arg\,min}

\title{Composing Ensembles of Pre-trained Models via Iterative Consensus}
\author{Shuang Li
\thanks{Correspondence to: Shuang Li $<$lishuang@mit.edu$>$.} \;\textsuperscript{\textdagger} \\ 
\thanks{indicates equal contribution.
Shuang Li did all the experiments on image generation, video question answering, and mathematical reasoning. Yilun Du did all the experiments on robot manipulation.}
MIT CSAIL \\
\texttt{lishuang@mit.edu} \\
\And
Yilun Du\textsuperscript{\textdagger} \\
MIT CSAIL \\
\texttt{yilundu@mit.edu} \\
\And
Joshua B. Tenenbaum \\
MIT CSAIL, BCS, CBMM \\
\texttt{jbt@mit.edu} \\
\AND
Antonio Torralba \\
MIT CSAIL \\
\texttt{torralba@mit.edu} \\
\And
Igor Mordatch \\
Google Brain \\
\texttt{imordatch@google.com} \\
}

\iclrfinalcopy % Uncomment for camera-ready version, but NOT for submission.
\begin{document}

\maketitle

\begin{abstract}
Large pre-trained models exhibit distinct and complementary capabilities dependent on the data they are trained on. Language models such as GPT-3 are capable of textual reasoning but cannot understand visual information, while vision models such as DALL-E can generate photorealistic photos but fail to understand complex language descriptions. In this work, we propose a unified framework for composing ensembles of different pre-trained models -- combining the strengths of each individual model to solve various multimodal problems in a zero-shot manner. We use pre-trained models as ``generators'' or ``scorers'' and compose them via closed-loop iterative consensus optimization. The generator constructs proposals and the scorers iteratively provide feedback to refine the generated result. Such closed-loop communication enables models to correct errors caused by other models, significantly boosting performance on downstream tasks, \eg improving accuracy on grade school math problems by $7.5\%$, without requiring any model finetuning. We demonstrate that consensus achieved by an ensemble of scorers outperforms the feedback of a single scorer, by leveraging the strengths of each expert model. Results show that the proposed method can be used as a general purpose framework for a wide range of zero-shot multimodal tasks, such as image generation, video question answering, mathematical reasoning, and robotic manipulation. Project page: \href{https://energy-based-model.github.io/composing-pretrained-models}{\texttt{https://energy-based-model.github.io/composing-pretrained-models}}.

\end{abstract}

\section{Introduction}
% Large pre-trained models have shown remarkable zero-shot generalization abilities, ranging from zero-shot image generation~\citep{ramesh2022hierarchical,saharia2022photorealistic} and natural language processing~\citep{brown2020language} to machine reasoning [cite] and action planning~\citep{ahn2022can,huang2022inner}. 
Large pre-trained models have shown remarkable zero-shot generalization abilities, ranging from zero-shot image generation and natural language processing to machine reasoning and action planning. 
Such models are trained on large datasets scoured from the internet, often consisting of billions of datapoints. Individual pre-trained models capture different aspects of knowledge on the internet, with language models (LMs) capturing textual information in news, articles, and Wikipedia pages, and visual-language models (VLMs) modeling the alignments between visual and textual information. 
While it is desirable to have a single sizable pre-trained model capturing all possible modalities of data on the internet, such a comprehensive model is challenging to obtain and maintain, requiring intensive memory, an enormous amount of energy, months of training time, and millions of dollars.
% While it is desirable to have a single large pre-trained model capturing all possible modalities of data on the internet, such a model is difficult to obtain, as even existing pre-trained models are expensive to train and memory intensive, costing millions of dollars to train consisting over billions of different individual parameters.
A more scalable alternative approach is to compose different pre-trained models together, leveraging
the knowledge from different expert models to solve complex multimodal tasks. 
% Furthermore, such an approach enables individual groups to focus on obtaining the best pre-trained models for a particular modalities, enabling the construction of scalable reasoning in a modular manner.

Building a unified framework for composing multiple models is challenging. Prior works~\citep{alayrac2022flamingo,zeng2022socratic} have explored composing pre-trained models in two main ways: (jointly) finetuning models on large datasets, or using common interfaces such as language to combine different models. 
However, these works have several key limitations:
First, simply combining models does not fully utilize each pre-trained model as there is no closed-loop feedback between models. Cascading models, such as Socratic models~\citep{zeng2022socratic}, allows one-way communication but prevents information processed by later models from propagating back to earlier models to correct errors. 
Secondly, common interfaces are limited to particular types of models. Language is used as the intermediate connection in Socratic models~\citep{zeng2022socratic}, but a language interface is insufficient to solve many real-world tasks, such as continuous robot control, which requires continuous representations. In addition, Socratic models require pre-designed language templates for the communication between models, which limits scalability.
Thirdly, jointly finetuning multiple models~\citep{alayrac2022flamingo} requires careful optimization to ensure that the model behaviors remain stable.
Such models also require intensive memory and large datasets and can only be used for solving specific tasks.

To resolve these difficulties, we propose a unified framework to compose models in a zero-shot manner\footnote{By zero-shot, we mean the composed models are never trained together on the evaluation task.} without any training/finetuning. 
Our framework employs a single model as a generator and an ensemble of scorers. 
% The generator iteratively constructs an output to achieve consensus from its scorers.
% Our framework defines a single model as a generator that iteratively constructs an output to achieve the consensus from an ensemble of scorer models. 
The generator iteratively generates proposals, and each scorer provides a feedback score indicating their agreement. 
The generator refines its outputs until all the scorers achieve a final consensus.
This iterative closed-loop communication between the generator and scorers enables models to correct the errors caused by other models, substantially boosting performance. 
% Feedback from each scorer is then used by the generator to further refine generated outputs iteratively, until finally a consensus output that is agreed upon every scorer is reached. 
% Results show that the ensemble of scorers improve the generator's performance through consistent interactions, outperforming the single feedback provided by scorers in the final stage.
% In addition, the ensemble scorers leverage crowd information when making decisions.

The ensemble of scorers is inspired by the idea of ``wisdom of the crowds’’. Each scorer provides complementary feedback to the generator, compensating for the potential weaknesses of other scorers. A Vision-Language scorer, for example, may correct the biases of a language model. We notice that different pre-trained model instances from the same family have diversity of outputs, which leads to more robust scorers. We demonstrate that guiding the generator with such an ensemble of scorers significantly outperforms a generator guided by a single scorer.

% Using a set of scorers enables the proposed method to leverage the ``wisdom of the crowds'' which may provide complementary feedback to the generator and compensate for the weaknesses of an individual model. A Vision-Language scorer, for example, may correct the biases of a language model. We illustrate that the application of an ensemble of scorers significantly outperforms the results made by a single scorer.

% By utilizing a set of scorers to jointly evaluate the output of a generator, we can directly leverage the ``wisdom of crowds'', and use the individual strengths of each separate scorer to compensate for weaknesses of other models. A Vision-Language scorer for example may correct for the textual biases of a language model. We illustrate how the application of an ensembles of models significantly improve zero-shot performance.

% multiple different scorers together, we can further leverage the ``wisdom of crowds'' and improve the performance over a single scorer. The ability to compose different models further enables our approach to be used in the continual learning setting where the underlying knowledge is dynamically changed over time. We may simply add new scorers on top of old scorers. Results show that models can be trained separately and new models and old models can be composed together to boost the results without any additional training or finetuning.

To summarize, our work has three main contributions.
\vspace{-5pt}
\begin{itemize}[leftmargin=1.2em]
\item First, we propose a unified framework for composing pre-trained models across a variety of tasks, such as image generation, video question answering, mathematical reasoning, and robot manipulation.
\item Second, we illustrate how the proposed framework can effectively solve zero-shot multimodal tasks without any training/finetuning.
The closed-loop communication between the generator and scorers allows the models to interact with each other to improve performance iteratively.
\item Finally, we illustrate how our framework enables the use of ensembles of different pre-trained models as scorers, significantly improving the zero-shot results by leveraging the strengths of multiple expert models.
% \item Finally, we illustrate how our framework enables the composition of separately trained scores and boost the performance by leveraging the crowd information when making decisions. The scorers can be learned at different time on different data, in a potentially incrementally learned manner, enabling the combination of incrementally learned knowledge.
\end{itemize}
\vspace{-5pt}
These observations point to the effectiveness of the proposed method as a general purpose framework for composing pre-trained models for solving various zero-shot multimodal tasks.

% \begin{figure}[t]
% \begin{center}
% \includegraphics[width=1\textwidth]{compose-foundation-models/figs/pdf/model.pdf}
% \end{center}
% \vspace{-10pt}
% \caption{\small \textbf{Overview.}
% }
% \label{fig:model_overview}
% \end{figure}

\begin{figure}[t]
\begin{center}
\includegraphics[width=1\textwidth]{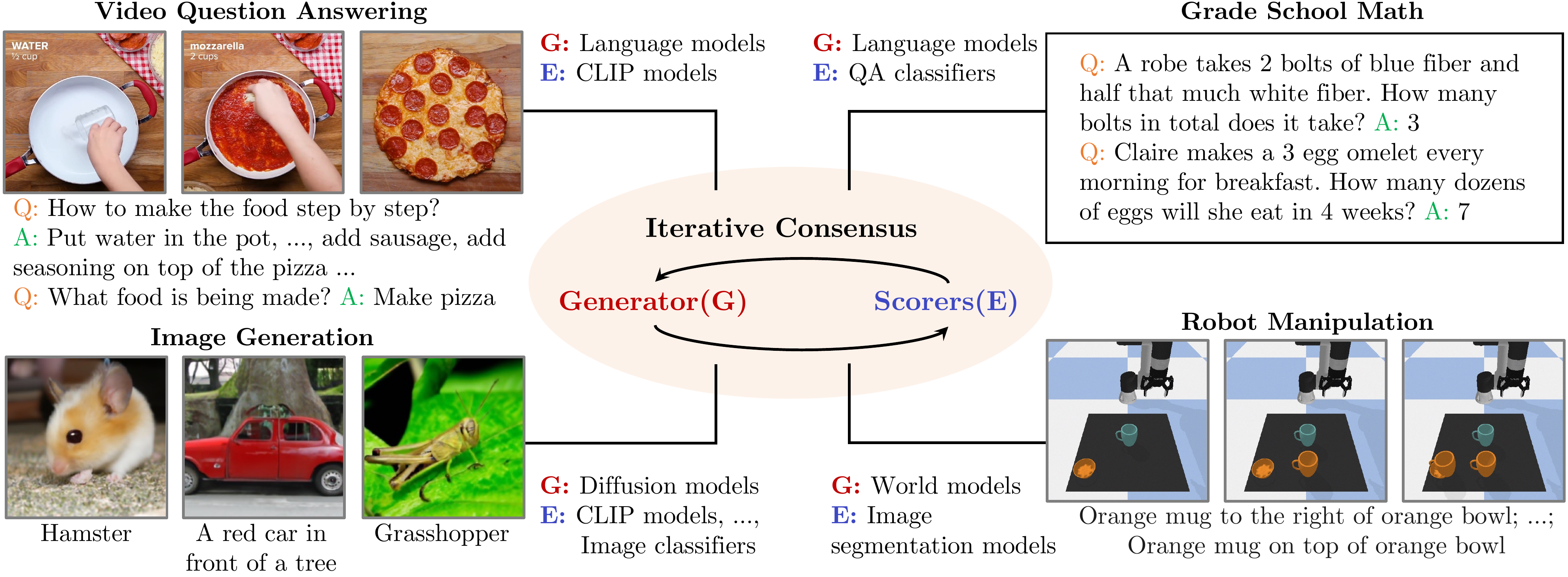}
\end{center}
\vspace{-5pt}
\caption{\small The proposed framework that composes a ``generator'' and an ensemble of ``scorers'' through iterative consensus enables zero-shot generalization across a variety of multimodal tasks.
}
\label{fig:model_overview}
\vspace{-10pt}
\end{figure}

\section{Related Work}
% \vspace{-3pt}
Large pre-trained models have shown great success across a variety of domains, such as language generation/translation, image generation, and decision-making.

\textbf{Language models.}
Large language models, such as ELMo~\citep{peters-etal-2018-deep}, BERT~\citep{devlin2018bert}, and GPT-2~\citep{radford2019language}, are able to achieve state-of-the-art performance on many standard NLP benchmarks. More recent works, such as GPT-3~\citep{brown2020language}, PALM~\citep{chowdhery2022palm}, and Chinchilla~\citep{hoffmann2022training} further enable few-shot learning from textual prompts. 
% In this paper, we explore the use of both GPT-2 and GPT-3 as large pretrained models of language information.
% exhibit few-shot learning ability, making them achievable for many real-world applications, such as question answering and chat bots.
% , including generating novels and talking with humans. 
% Given proper language prompts, GPT-3 can generate sentences that are hard to be distinguished by humans.

% Joint training across Modalities
% Multimodal Probabilistic Models
% and Prompting

\textbf{Vision-language models.}
Large pre-trained vision-language generative models, such as DALL-E 2~\citep{ramesh2022hierarchical}, Parti~\citep{yu2022scaling}, and Imagen~\citep{saharia2022photorealistic}, can generate high-resolution images given natural language descriptions. Large pre-trained vision-language discriminative models, such as CLIP~\citep{radford2021learning}, convert images and languages into the same feature space, achieving remarkable zero-shot generalization ability on downstream tasks.

% can reliably caption images and achieving remarkable zero-shot classification performance. 
% In this paper, we explore the use of GLIDE, as a large pretrained generative model, and CLIP as a large pretrained discriminative model.

% Most of these models~\citep{ramesh2022hierarchical,saharia2022photorealistic} are built on Diffusion Models~\citep{ho2020denoising,sohl2015deep}.
% for its tractability and flexibility. 
% Stable Diffusion~\citep{rombach2022high}, accelerates the image generation speed by performing the diffusion process in the latent space rather than the pixel space.
% Other vision-language models, .

% video-language

\textbf{Decision-making models.}
Large pre-trained models have been widely applied to solve decision-making tasks, such as learning general purpose policies~\citep{reed2022generalist,li2022pre,shridhar2022cliport}, making planners~\citep{huang2022inner, ahn2022can}, and learning world models~\citep{ebert2018visual}. 
However, due to the large variability in decision-making tasks, no existing pre-trained models can be readily applied across different tasks. 
% In this paper, we explore the idea of having a hypothetical large pretrained generative model of the world \citep{lowrey2018plan}.
% Many recent works~\citep{reed2022generalist,shridhar2022cliport,baker2022video,huang2022language} use large pre-trained models for decision-making.
% Gato~\citep{reed2022generalist} and LID~\citep{li2022pre} propose general-purpose frameworks for solving multimodal tasks across a variety of environments. 
% VPT~\citep{baker2022video} plays Minecraft games by learning from massive online videos.
% More recent works, InnerMonlogue~\citep{huang2022inner} and SayCan~\citep{ahn2022can}, enable zero-shot robot planning through pre-trained language models. 
% VPT~\citep{baker2022video} shows remarkable performance in playing Minecraft games using policies trained on online videos.
% In this work, we propose a unified framework that can compose various pre-trained models for solving language, vision, and decision-making tasks.

\textbf{Composing pre-trained models.} 
Composing large pre-trained models has been widely studied recently. 
The predominant way to compose pre-trained models is to (joint) finetune them on new tasks~\citep{li2019visualbert,wang2021simvlm,alayrac2022flamingo,mokady2021clipcap}, but such approaches are computationally expensive. 
Alternative approaches compose models through a common interface such as language\citep{tewel2021zero,zeng2022socratic}.
Other works compose pre-trained models by composing learned probability distributions of the data, such as energy-based models~\citep{liu2022compositional,liu2021learning,du2020compositional}, which can be applied to image generation.
% Other works compose pre-trained models by composing the data probability distributions they modeled, such as energy-based models~\citep{liu2022compositional,liu2021learning,du2020compositional}, which can be composed to generate new combinations of images.
% combine the probability distributions modeled by different models, for example composing visual generation using energy-based models~\citep{liu2022compositional,liu2021learning,du2020compositional}.
% that compose energy-based models during inference for visual generation.
%  probabilistic without training also include 
In this paper, we propose a general framework to compose pre-trained models across a variety of domains without any training or finetuning.

\section{Method}
% \vspace{-3pt}
% Composing Foundation \textbf{M}odels through \textbf{IT}erative Refinement (MIT)

Given a set of large pre-trained models, we aim to utilize the expert knowledge from different models to solve zero-shot multimodal tasks. We separate pre-trained models into two categories -- generators ($G$) such as GPT~\citep{brown2020language,radford2019language} and Diffusion models~\citep{ho2020denoising} that can generate candidate solutions, and scorers ($E$) such as CLIP~\citep{radford2021learning} and classifiers that output a scalar score to evaluate each generated solution.
We propose \textbf{\modelname} (composing ensembles of \textbf{P}re-trained models via \textbf{I}terative \textbf{C}onsensus), a framework which composes ensembles of pre-trained models for multimodal tasks.
% Given a set of large pre-trained models, we aim to utilize the expert knowledge from different models to solve zero-shot multimodal tasks. We categorize pre-trained models into two separate categories -- generators, such as GPT~\citep{brown2020language,radford2019language} or Diffusion models~\citep{ho2020denoising,sohl2015deep}, which we refer to as $G$ are responsible for generating candidate solutions, and energy scores, which we refer to as $E_i(\vx): \mathcal{R}^n \rightarrow \mathcal{R}$, such as CLIP and classifiers, which are discriminative models that are responsible for assessing and scoring solutions. 
% compose different models together jointly to generate solution through an iterative energy minimization procedure. 
The core idea of \modelname is to generate solutions through iterative optimization, where we leverage the knowledge from different models to jointly construct a consensus solution.
In \modelname, a generator $G$ iteratively and sequentially generate candidate solutions, each of which is refined based on the feedback from a set of scorers. In particular, we seek to obtain a solution $x^{*}$ such that
% \vspace{-3pt}
\begin{equation}
    x^{*} = \argmin_{x \sim G} \sum_n E_n(x),
    \label{eq:compose_energy}
% \vspace{-1pt}
\end{equation}
where $\{E_n\}$ is the set of scorers. At each iteration, we refine the solutions to have a lower score than the previous iterations. This procedure, described in
\eqn{eq:compose_energy}, converges to a solution that minimizes the energy across multiple pre-trained models, which maximizes the agreement between the generator and scorers.
In contrast to Socratic Models where different pre-trained models are called sequentially, the closed-loop iterative refinement through which we obtain $x^{*}$ enables the generator and scorers to communicate with each other to reach a consensus on the final solution.

% is satisfied across different scorers $E_i(\vx)$ -- each of which assess the validity of solution using different aspects of captured knowledge in the given model.
% The iterative procedure through which a minimal energy solution found across different pre-trained models serves as a closed loop feedback approach to obtain solutions from pre-trained models. 
% In contrast to approaches, such as Socratic Models, where outputs from different pre-trained models and sequentially nested, the iterative refinement through which we obtain $\vx^{*}$ enables all models, both the generators and scorers, to inform the final solution in a closed-loop fashion.

Below, we illustrate how \modelname can be broadly applied across tasks in image generation, video question answering, grade school math, and robot manipulation. To optimize \eqn{eq:compose_energy}, we consider two different optimization procedures -- either a continuous approach that leverages the gradients of each scorer $E_n(x)$ or a discrete approach that directly samples possible solutions.

\begin{figure}[t]
\begin{center}
\includegraphics[width=1\textwidth]{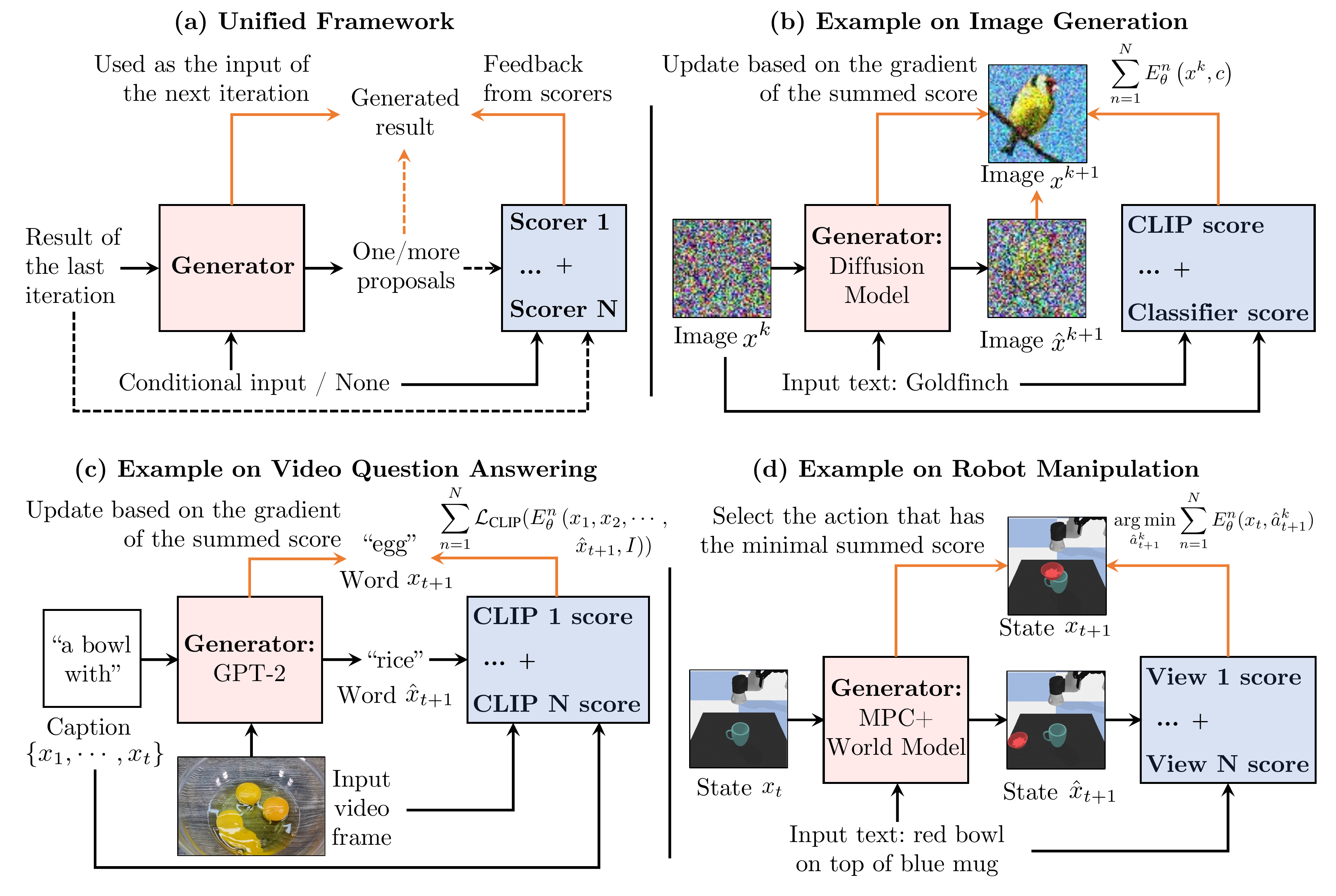}
\end{center}
\vspace{-5pt}
\caption{\small \textbf{The proposed unified framework and examples on three representative tasks.} (a) Overview of the proposed unified framework. Dashed lines are omitted for certain tasks. (b) Image generation. A pre-trained diffusion model is used as the generator, and multiple scorers, such as CLIP and image classifiers, are used to provide feedback to the generator.
(c) Video question answering. GPT-2 is used as the generator, and a set of CLIP models are used as scorers. 
(d) Robot manipulation. MPC+World model is used as the generator, and a pre-trained image segmentation model is used to compute the scores from multiple camera views to select the best action. 
Orange lines represent the components used to refine the generated result. 
}
\label{fig:model_details1}
% \vspace{-5pt}
\end{figure}

% 1. compose generator and energy scorer
% 2. compose multiple energy scorers

% We propose MIT, a framework that composes foundation \textbf{M}odels through \textbf{IT}erative refinement.
% Our framework consists of a generator, one or multiple energy scorers. 
% The generator generates a proposal $x^{k}$ while the energy scorer provides feedback on the proposal through the gradient $\nabla_{x} E_{\theta}\left(x^{k}\right)$. The iterative refinement is an MCMC sampling which can be written as follows:
% \begin{equation}
%     x^{k+1}=x^{k}-\frac{\lambda}{2} \nabla_{x} E_{\theta}\left(x^{k}, c \right),
% \end{equation}

% \begin{equation}
%     \sum_{n=1}^N \nabla_{x} E_{\theta}^n\left(x^{k}, c \right)
% \end{equation}

% \begin{equation}
%     \argmin \sum_k E_{\theta}\left(x^{k}, c \right)
% \end{equation}

% where $k$ is the iteration step. $E_{\theta}$ can be any function that provides an energy value to evaluate the proposal $x^{k}$. 

% We propose an elegant way to compose different types of foundation models. Some existing methods can be reformulated into our framework as well, such as vqgan+clip and diffusion+clip, where the vqgan and diffusion models are the generator and the clip is the energy scorer. 

\subsection{Applications to Zero-shot Tasks}
\label{sec:method_tasks}
% \vspace{-3pt}
\textbf{Image generation.}
We first apply the proposed framework to image generation to generate images conditioned on a text description or a class label. 
% There is one image generator and multiple scorers that can provide feedback for the generated image. 
We use the reverse diffusion process of GLIDE~\citep{nichol2021glide}, a text-guided diffusion model, as the generator to generate image proposals. 
At each step of the diffusion process (corresponding to a step of the iterative refinement), we use the gradient from an ensemble of scorers, such as CLIP~\citep{radford2021learning}, to guide and update the generated proposals. We iteratively repeat this procedure until the final step.
% , text-image classifiers~\citep{dhariwal2021diffusion}, and the classifier-free guidance~\citep{ho2022classifier}.

% , to provide feedback for the generator to refine the generated result. 
As shown in \cref{fig:model_details1} (b), the image $x^{k}$ generated at iteration $k$ is first sent to the diffusion model to generate an image proposal $\hat{x}^{k+1}$. Each scorer outputs a score to evaluate whether the generated image matches the given text input. 
For example, CLIP computes the cosine similarity between the image and text features as the score. 
% The image classifier predicts the probability of the image matching the text label. The classifier-free guidance can be treated as an implicit classifier that directly provides pixel-wise gradient feedback to the generated image.
The scores generated by different scorers are summed, and their gradient with respect to $x^{k}$ is used to compute the next reverse prediction $x^{k+1}$:
% \vspace{-3pt}
\begin{equation}
    % x_{t+1}=x_t-\frac{\lambda}{2} \nabla_{x} \sum_{n=1}^N E_{\theta}^n \left(x_{t}, c \right),
    % x_{t+1} =\mathcal{N}(\hat{x}_{t+1} + \lambda \nabla_{x} \sum_{n=1}^N E_{\theta}^n \left(x_{t}, c \right), \sigma^2),
    x^{k+1} \leftarrow \hat{x}^{k+1} + \lambda \nabla_{x^k} \sum_{n=1}^N E_{\theta}^n \left(x^{k}, c \right),
% \vspace{-1pt}
\end{equation}
% where $\mathcal{N}$ is the normal distribution, 
where $N$ is the number of scorers and $c$ is the text label. We denote the reverse process prediction as $x^{k+1}$ instead of $x^{k-1}$ (used by most diffusion models) to keep the consistent notation across tasks.
% and $\sigma^2$ is the variance. 

\textbf{Video question answering (VQA).}
Caption generation for a single video frame is shown in \cref{fig:model_details1} (c). We use GPT-2 as the generator and multiple different CLIP models, trained with different configurations, as the scorers. 
%  use GPT-2 to summarize the captions and answer questions, and as our scorers, we utilize multiple different CLIP models, trained with different configurations for zero-shot video frame captioning.
%
% Given a video frame $I$ and a text prompt, such as ``Image of'', 
Given a video frame $I$, we generate a sequence of words to describe it. To integrate feedback from scorers to the generator, similar to \citep{tewel2021zero}, we define a context cache $C_t$ (a set of embedding functions in GPT-2) that stores the context information generated so far, which is updated iteratively based on the feedback from scorers. 
The prediction of the next word from the generator $G$ is given by $x_{t+1}=G(x_t, C_t)$. 
% The goal is to update $C_t$ iteratively based on the CLIP score to generate the next word such that the sentence is grammatically sound as well as accurately describes the given video frame. 
To update $C_t$, we first use $G$ to generate a set of candidate words $\hat{X}_{t+1} = \{\hat{x}_{t+1}\}$, and then use the feature distance (after softmax) between each sentence (the concatenation of previous words and each new word $\{x_1, x_2, \cdots, \hat{x}_{t+1}\}$, where $\hat{x}_{t+1} \in \hat{X}_{t+1}$) and the video frame as the probability of them matching. The CLIP score is the cross-entropy loss $\mathcal{L_{\text{CLIP}}}$ between this new probability distribution and the original distribution of the next word obtained from the generator $G$.
% The history tokens $\{x_1, \cdots, x_t\}$ is first sent to the generator to predict the next token $\hat{x}_{t+1}$.
% Then the scorers compute the feature distances (scores) between the new sentence (concatenation of history tokens and the new token) and the given video frame.
% Similar to image generation, 
The gradient of the summed score (multiple CLIP models) is then propagated to $G$ to update $C_t$:
% \vspace{-3pt}
\begin{equation}
    C_{t}^{k+1} \leftarrow C_{t}^k + \lambda \nabla_{C_{t}^k} \sum_{n=1}^N \mathcal{L_{\text{CLIP}}} (E_{\theta}^n \left( x_1, x_2, \cdots, \hat{x}_{t+1}, I \right)),
% \vspace{-3pt}
\label{eqn:update_c}
\end{equation}
where $k$ is the step of iterative refinement.
After several iterations, the updated $C_t$ is used to generate the next token $x_{t+1}=G(x_t, C_t)$. We repeat this process until we generate the entire caption.
We cascade the captions of multiple video frames and questions about this video to prompt GPT-3 for video question answering (See \cref{apx_sec:vqa}).

% We first use \modelname to generate video frame captions. We then use GPT-3 to summarize the captions and answer questions about this video.

% Results show that utilizing the proposed framework and GPT-3 enables effective video question answering.

% \begin{equation}
%     \sum_{n=1}^N E_{\theta}^n \left(x_1, \cdots, \hat{x}_{t+1}, c \right),
% \end{equation}

\textbf{Grade school math.}
We further apply \modelname to solve grade school math problems. We use GPT-2 as the generator and treat the grade school math problem as a text generation problem. The scorer, a pre-trained question-solution classifier, provides the generator feedback to guide the next token's generation $x_{t+1}$. We follow the approach used in VQA to iteratively optimize the generations based on the feedback from scorers. Our generator $G$ first generates a set of candidate words $\hat{X}_{t+1} = \{\hat{x}_{t+1}\}$, and then the classifier predicts the probability of each solution (the concatenation of previous words and each new word $\{x_1, x_2, \cdots, \hat{x}_{t+1}\}$, where $\hat{x}_{t+1} \in \hat{X}_{t+1}$) matching the given question. 
The classifier score is the cross-entropy loss between this new probability distribution and the original distribution of the next word obtained from the generator $G$. The gradient of the classifier score is used to update $C_t$ through iterative refinement, same as \cref{eqn:update_c}. The updated $C_t$ is used to predict the next word $x_{t+1}=G(x_t, C_t)$. We repeat this process until we generate the complete solution.

% \begin{equation}
%     C_{t}^{k+1} \leftarrow C_{t}^k + \lambda \nabla_{C_{t}^k} \sum_{n=1}^N \mathcal{L_{\text{Classifier}}} (E_{\theta}^n \left( x_1, x_2, \cdots, \hat{x}_{t+1}, Q \right)),
% % \vspace{-3pt}
% \label{eqn:update_c_math}
% \end{equation}

% The pre-trained scorer is a classifier to evaluate the correctness of the output answer for the given math problem.

% \input{compose-foundation-models/figs/model_details2}

\textbf{Robot manipulation.} Finally, we illustrate how \modelname can be applied to manipulate objects in the robot environment to conform to a set of object relations such as ``red bowl on top of blue mug'' shown in \cref{fig:model_details1} (d).
% or real-world images that illustrate the relations as shown in \cref{fig:model_details1}.
We use the combination of Model Predictive Control (MPC) \citep{williams2015model} and the World Model as the generator.
At each time step, we first use MPC to sample a set of possible actions and then render the state images (after executing an action) from multiple camera views using the world model. For each action, the scorer computes a summed score across all camera views as its final score, which is used to select the best action to execute. Thus, in this domain, the ensemble consists of scorers based on different views of the scene.

For the generator, we assume that there is a pre-trained model, \ie world model, that can accurately render and simulate the dynamic changes in the robot world. Since such a large pre-trained model does not directly exist, we approximate it using an environment simulator combined with MPC as the generator.
For the scorer, we use the pre-trained ViLD~\citep{gu2021open} to generate segmentation maps for images captured by different camera views $n$, and the corresponding text label for each segment, which are used to obtain object relations.
We compare the generated object relations and the relations specified by the text description to obtain the score, \ie score equals 0 if they match; otherwise, 1 (here the score means the distance) (see \cref{apx_sec:robot} for details). 
%
% To obtain a final world state $x_T$ that satisfies the specified relations, and the sequence of actions $\{a_1,\cdots, a_T\}$ that manipulate the objects into the final state $x_T$, we iteratively update the state of the world $x_t$ to optimize scorers using the following rule:
% \begin{equation}
%     x_{t+1} \leftarrow x_{t} + \argmin_{\hat{a}_{t+1}} \sum_{n=1}^N E_\theta^n(x_t, \hat{a}_{t+1}).
% \vspace{-3pt}
% \end{equation}
% Each scorer, $E_\theta^n$, outputs a score for the resultant state obtained when a candidate action $\hat{a}_{t+1}$ is applied to the current world state $x_t$. 
%
To obtain a final world state $x_T$ that satisfies the specified relations, and the action sequence $\{a_1,\cdots, a_T\}$ that manipulates the objects into the final state $x_T$, the generator iteratively samples possible actions $\hat{a}_{t+1}^k$ and gets feedback from scorers. The best action is selected as:
% \vspace{-5pt}
\begin{equation}
    a_{t+1} = \argmin_{\hat{a}_{t+1}^k} \sum_{n=1}^N E_\theta^n(x_t, \hat{a}_{t+1}^k).
% \vspace{-5pt}
\end{equation}
Each scorer, $E_\theta^n$, outputs a score for the resultant state obtained when a candidate action $\hat{a}_{t+1}^k$ is applied to the current world state $x_t$. We execute $a_{t+1}$ in the environment and get a new state $x_{t+1}$. We repeat this process until the task is accomplished or we are at the final step $T$.

\section{Experiment Setup}
% \vspace{-3pt}
We evaluate the proposed framework for composing pre-trained models on four representative tasks, including image generation, video question answering, grade school math, and robot manipulation.

\textbf{Image generation.}
% We first show that composing the pre-trained image generation model, \ie GLIDE, and multiple scorer models, \ie CLIP, image classifier, and classifier-free guidance, enables effective zero-shot image generation. 
We first show that composing the pre-trained image generator and scorer models such as CLIP enables effective zero-shot image generation. 
We evaluate the image generation results on ImageNet~\citep{deng2009imagenet} with the image resolution of $64\times64$. The class labels are used as the text input to guide image generation. Each method generates 50 images for each class.
We evaluate the image generation quality using Inception Score (IS)~\citep{salimans2016improved}, Fréchet Inception Distance (FID)~\citep{heusel2017gans}, and Kernel Inception Distance (KID)~\citep{binkowski2018demystifying}.
% to evaluate approaches from different perspectives. 
IS measures the distribution of generated images. Higher values mean the models can generate more distinct images. FID considers the distributions of both generated images and real images. Lower scores represent that the generated images are closer to the real images. KID is similar to FID, measuring the similarity between two data distributions, but is in the kernel space.

\textbf{Video question answering.}
We evaluate methods for solving VQA tasks on ActivityNet-QA~\citep{yu2019activitynet}.
Our method generates free-form language answers instead of selecting an answer from a pre-defined answer set~\citep{yang2021justask,lei2022revealing}. To evaluate such free-form VQA, we ask workers from Amazon Mechanical Turk to measure whether the generated answer matches the given question and video (See \cref{apx:mturk_exp} for IRB approval and experimental details).
For fair comparisons, all the approaches answer the same 300 video questions, and each answer is evaluated by three different workers.
The accuracy rate and vocabulary size are reported. 
An answer is correct if at least two workers believe it is correct. The accuracy rate is the percentage of correctly answered questions over all the questions. 
To evaluate the diversity of generated answers, we also report the vocabulary size (\ie the number of words) of answers generated by each method.

\begin{table*}[t]
\small
  \caption{\small{ \textbf{Image generation results on ImageNet.} 
%   G is the generator and E means energy scorers. 
  Our \modelname can compose the pre-trained generator (G) and scorers (E) through iterative optimization. Composing multiple scorers further boosts performance.}}
  \vspace{-5pt}
  \label{exp:image_generation}
  \centering
  \scalebox{0.8}{
  \begin{tabular}{lllcccccc}
    \toprule
    \bf Method Name & \bf Generator & \bf Scorer & \bf IS $\uparrow$  & \bf FID $\downarrow$ & \bf KID $\downarrow$ \\
    \midrule
    \bf \modelname (G+E1) & GLIDE & CLIP & 25.017 & 30.462 & 6.174 \\
    \bf \modelname (G+E2) & GLIDE & CLS & 22.077 & 30.871 & 7.952 \\
    \bf \modelname (G+E3) & GLIDE & CLS-FREE & 25.926 & 29.219 & 5.325 \\
    \midrule
    \bf \modelname (G+E1+E2+E3) & GLIDE & CLIP + CLS + CLS-FREE & \bf 34.952 & \bf 29.184 & \bf 3.766 \\
    \bottomrule
  \end{tabular}
  }
  \vspace{-5pt}
\end{table*}

\begin{table*}[t]
\small
  \caption{\small{\textbf{Video question answering results on ActivityNet-QA.} JustAsk (FT) is finetuned on ActivityNet-QA, thus achieving the best results. For zero-shot VQA, our method (\modelname) significantly outperforms JustAsk (Pretrain), one of the best VQA methods. Using multiple scorers further improves the performance.}}
  \vspace{-5pt}
  \label{exp:vqa}
  \centering
  \scalebox{0.85}{
  \begin{tabular}{llllccccc}
    \toprule
    \bf Method Name & \bf Zero-Shot & \bf Generator & \bf Scorer & \bf Accuracy $\uparrow$ & \bf Vocab $\uparrow$ \\
    \midrule
    \bf JustAsk (FT) & No & - & - & \bf 64.667 & 160 \\
    \midrule
    \bf JustAsk (Pretrain) & Yes & - & - & 50.671 & 210 \\
    % \midrule
    \bf \modelname (G+E1) & Yes & GPT-2 & CLIP-32 & 58.389 & 267 \\
    \bf \modelname (G+E1+E2+E3) & Yes & GPT-2 & CLIP-32 + CLIP-14 + CLIP-multilingual & \bf 61.168 & \bf 304 \\
    \bottomrule
  \end{tabular}
  }
  \vspace{-10pt}
\end{table*}

\begin{figure}[t]
\begin{center}
\includegraphics[width=1\textwidth]{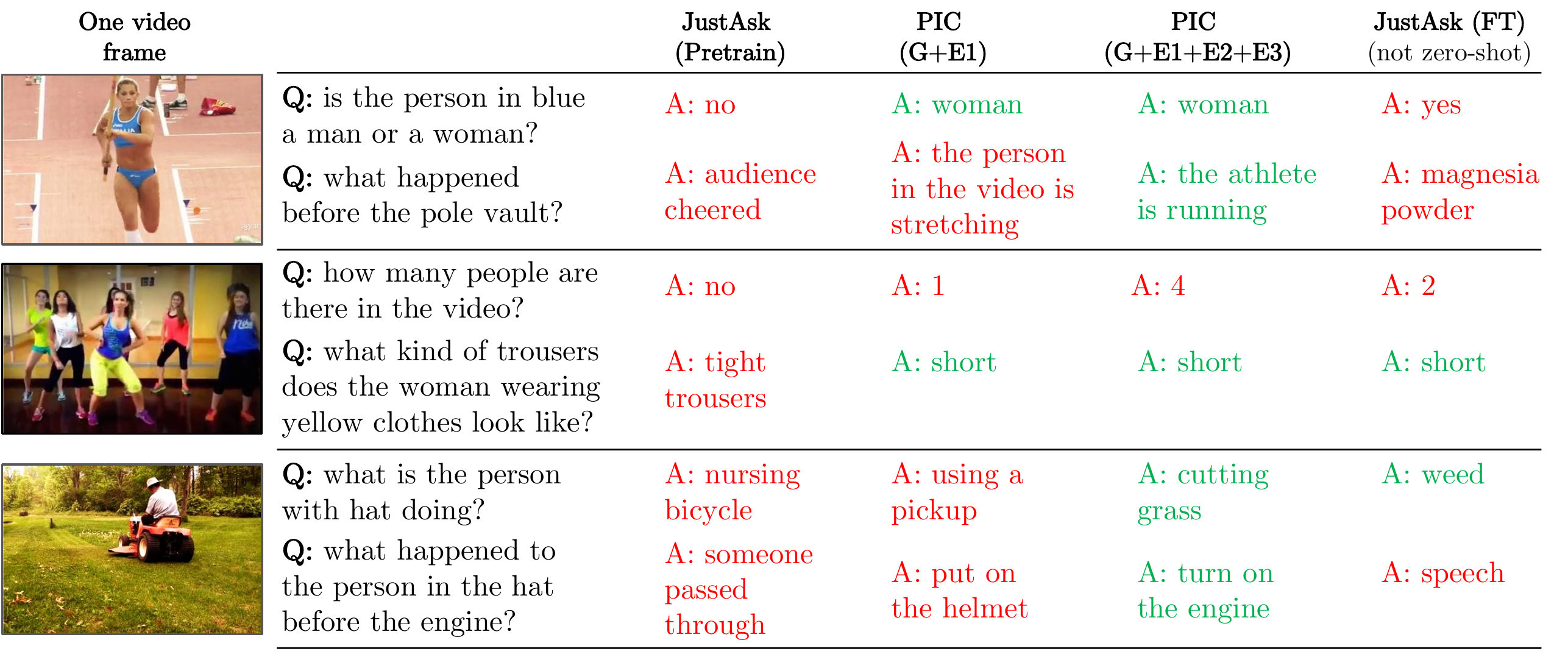}
\end{center}
\vspace{-10pt}
\caption{\small \textbf{Video question answering example results.} Our approach successfully identifies gender and clothing, but its failure to count objects is a reflection of GPT-2 and CLIP’s inability to count.
}
\label{fig:vqa_result}
\vspace{-5pt}
\end{figure}

\textbf{Grade school math.}
GSM8K~\citep{cobbe2021training} is a dataset for grade school math problems.
Each problem consists of a question, intermediate analyses, and a final solution.
We evaluate approaches to solving problems on the 1K test set. We use beam search to generate candidate solutions. 
The accuracy of beam size 1 and beam size 5 are reported.
For beam size of 1, we mark the result as correct if it matches the final solution.
For beam size of 5, we mark the result as correct if any of the five generated results matches the solution.

\textbf{Robot manipulation.} We next evaluate how pre-trained models may be used to manipulate objects in Ravens~\citep{zeng2020transporter}. 
In Ravens, the action space of robot is to drop an object at a 2D location on the table. The goal is to obtain a scene configuration that satisfies the object relations specified by a textual description or a real-world image, such as ``blue mug to the left of purple bowl''. 
The task is successful if the object relations in the final state satisfy all the relations specified by the input text or image.
We report the success rate of tasks with two and three specified object relations.

% 100 language goal
% 10 image goal

% We measure the relation accuracy and overall accuracy in the final state. 
% Relation accuracy means the object relations satisfy the relations specified by the input text or image. Overall accuracy means both the objects and their relations are correct. 

% We measure the relation accuracy and overall accuracy in the final state. 
% Relation accuracy means the object relations satisfy the relations specified by the input text or image. Overall accuracy means both the objects and their relations are correct. 
% We report the accuracy on two object relations and three object relations.

% into a target set of relations, either specified through text, or specified from real images. At each timestep, the action space of the robot is to drop an object at a 2D location in a plane. After robotic manipulation, we measure the accuracy of the final state in having the relations specified, though not necessarily the correct objects (relation accuracy), and overall accuracy, having both correct objects and relations between the objects. 

% N
% To manipulate a scene subject to a set of relations, we then utilize a MPC procedure to directly minimize \eqn{eq:compose_energy}, where we sample different possible actions in the environment and utilize our generator G to render different viewpoints of the resultant scene that our scored by our seperate scorers. The action with maximal score with respect to scorers is then taken.

% \vspace{-3pt}
\section{Experiments}
% \vspace{-3pt}
We compare the proposed method with baselines on the above four zero-shot tasks.
% from \cref{expsec:image_generation} to \cref{expsec:grade_school_math}.

%% ---------------------------------------------------------------------
% \vspace{-3pt}
\subsection{Image Generation}
\label{expsec:image_generation}
% \vspace{-3pt}
We evaluate the zero-shot conditional image generation on ImageNet in \cref{exp:image_generation}. We first show results of composing a single generator (G) and a single scorer (E). We compose GLIDE~\citep{nichol2021glide} with three different types of scorers, respectively.
% , \ie \emph{G+E1}, \emph{G+E2}, and \emph{G+E3}.
% E1 is the CLIP model~\citep{radford2021learning}, E2 is the image classifier~\citep{dhariwal2021diffusion}, and E3 is the classifier-free guidance~\citep{ho2022classifier}.
E1 is CLIP~\citep{radford2021learning} that computes the cosine similarity between the image and text features as the score, E2 is the image classifier (CLS)~\citep{dhariwal2021diffusion} that predicts the probability of the image matching the text label as the score, and E3 is the classifier-free guidance (CLS-FREE)~\citep{ho2022classifier} which can be treated as an implicit classifier that directly provides pixel-wise gradient feedback to the generated image (\cref{apx_sec:image_generation}).
We then compose the generator with all scorers, \ie G+E1+E2+E3. 
Composing the generator and a single scorer allows zero-shot image generation. Composing multiple scorers significantly outperforms a single scorer. We note that the generator is not trained on ImageNet; thus the results in \cref{exp:image_generation} cannot be directly compared with methods trained on ImageNet.

%% ---------------------------------------------------------------------

% \vspace{-3pt}
\subsection{Video question answering}
\label{expsec:vqa}
% \vspace{-3pt}

\textbf{Quantitative results.}
We compare \modelname with one of the state-of-the-art VQA approaches, \ie JustAsk~\citep{yang2021justask}, on ActivityNet-QA~\citep{yu2019activitynet}.
In \cref{exp:vqa}, JustAsk (FT) is finetuned on ActivityNet-QA, thus achieving the best results.
We then compare \modelname with JustAsk (Pretrain) for zero-shot VQA. 
The generator of our method, GPT-2 (medium size), is trained on Webtext~\citep{radford2019language} using the Huggingface library~\citep{wolf2019huggingface}.
Our scorers are CLIP models~\citep{radford2021learning,reimers-2019-sentence-bert} trained on different datasets or using different configurations.
\modelname (G+E1) outperforms JustAsk (Pretrain) by $\%7.72$. Composing more scorers further improves the accuracy by $\% 2.78$. In addition, the vocabulary size of answers generated by our method is larger than other approaches, indicating that our method can answer questions using richer language and more diverse phrasing.
Note that our method solves a ``more challenging'' problem than JustAsk (Pretrain) and JustAsk (FT).
Our method generates open-language answers while JustAsk (Pretrain) and JustAsk (FT) select an answer from a pre-defined answer set. 
Generating free-form responses requires both semantic and grammatical correctness. \modelname performs well on both these dimensions while also using a richer vocabulary.

% Generating open-language answers requires both the correctness of the language itself and the answer matching the question. The strong performance and the ability to create richer vocabulary further show the effectiveness of the proposed method in solving zero-shot VQA tasks.

\textbf{Qualitative results.}
In \cref{fig:vqa_result}, we show answers generated by different approaches given a video (only showing a single video frame) and questions.
Our approach successfully identifies gender and clothing, but none of the approaches know how to count numbers.

\subsection{Grade school math}
\label{expsec:grade_school_math}
% \vspace{-5pt}

\textbf{Quantitative results.}
In \cref{exp:grade_school_math}, we compare \modelname with two baselines, \ie GPT-Pretrain and GPT-FT, for solving math problems on GSM8K~\citep{cobbe2021training}. 
GPT-Pretrain uses the pre-trained GPT-2 (medium size GPT-2 trained on Webtext using Huggingface) to generate numeric strings.
GPT-FT is based on GPT-Pretrain and then finetuned on GSM8K. 
Our method uses the same GPT-2 (Pretrain) as the generator and a question-solution classifier (CLS) as the scorer. The classifier is trained on GSM8K to distinguish whether a solution is correct for a given question.

\begin{wraptable}{r}{7.6cm}
\small
  \centering
  \vspace{-10pt}
  \caption{\small{ \textbf{Grade school math results on GSM8K.} Our method (\modelname) that composes GPT-2 and a pre-trained question-solution classifier significantly outperforms the baselines, including GPT-FT that is finetuned on GSM8K.}}
  \vspace{-5pt}
  \label{exp:grade_school_math}

  \scalebox{0.83}{
  \begin{tabular}{lllcccccc}
    \toprule
    \bf Method Name & \bf Generator & \bf Scorer &  \bf BS=1 $\uparrow$ & \bf BS=5 $\uparrow$  \\
    \midrule
     \bf GPT-Pretrain & GPT-2 (Pretrain) & - & 1.744 & 12.206 \\
     \bf GPT-FT & GPT-2 (FT) & - & 3.487 & 18.271  \\
     \midrule
     \bf \modelname (G+E) & GPT-2 (Pretrain) & CLS & \bf 16.831 & \bf 20.773 \\
    \bottomrule
  \end{tabular}
  }
  \vspace{-5pt}
\end{wraptable}

We surprisingly find that \modelname achieves significantly better performance than GPT-FT (\%13.344 higher on beam size 1), even though the generator has never seen the math problems before. The classifier only provides feedback to the generator, but through iterative refinement, combining a generator and a scorer without joint training is more effective than directly finetuning GPT-2 on GSM8K (we find the overfitting problem when finetuning GPT-2 on GSM8K).

\begin{figure}[t]
\begin{center}
\includegraphics[width=1\textwidth]{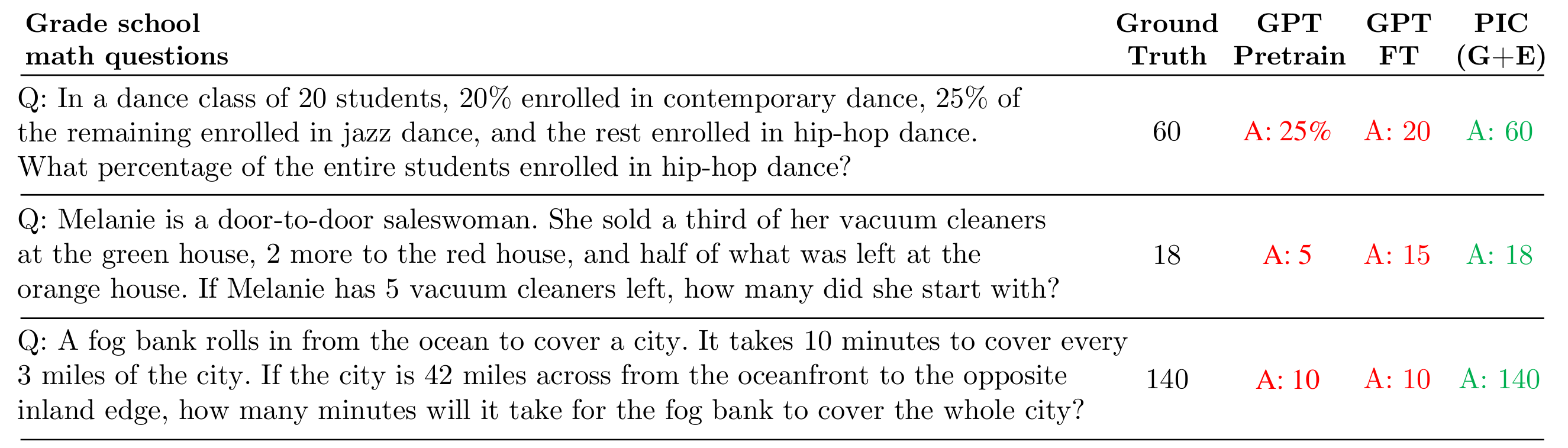}
\end{center}
\vspace{-10pt}
\caption{\small \textbf{Grade school math example results.}  Our method can solve math problems involving addition, subtraction, multiplication, and division.
}
\label{fig:math_result}
% \vspace{-5pt}
\end{figure}

\textbf{Qualitative results.}
Example results of different methods are shown in \cref{fig:math_result}.
Our method can solve math problems involving addition, subtraction, multiplication, and division, even for solutions with three-digit numbers. In contrast, GPT-FT often fails to understand math problems.

\begin{wraptable}{r}{6.6cm}
\small
\vspace{-15pt}
  \caption{\small{ \textbf{Robot manipulation results on Ravens.} \modelname can manipulate objects to achieve object relations specified by textual descriptions (Text) or real-world images (Image). Using scorers of multiple camera views substantially improves the success rate.
  }}
  \vspace{-5pt}
  \label{exp:robot_main}
  \centering
  \setlength{\tabcolsep}{0.35em}
  \scalebox{0.9}{
  \begin{tabular}{l|cc|ccccccc}
    \toprule
    \bf Method Name  &  \multicolumn{2}{c}{\bf 2 Relations} &  \multicolumn{2}{|c}{\bf 3 Relations} \\
    & Text $\uparrow$ & Image $\uparrow$ & Text $\uparrow$ & Image $\uparrow$  \\
    \midrule
    % 45.0
     \bf \modelname (G+E1)  & 35.0 & 27.5  & 50.0  &  45.0 \\
     \bf \modelname (G+$\bf \sum_{n=1}^5 E_n$) & \bf 67.5 & \bf 52.6 & \bf 75.0 & \bf 65.3 \\
    \bottomrule
  \end{tabular}
  }
  \vspace{-5pt}
\end{wraptable}

% \vspace{-3pt}
\subsection{Robot manipulation}
\label{expsec:robot}
% \vspace{-3pt}

\textbf{Quantitative results.}
We evaluate the proposed method of manipulating objects to achieve object relations specified by the textual descriptions (Text) or real-world images (Image).
In \tbl{exp:robot_main}, we find that using scorers of multiple camera views substantially improves the accuracy on both settings.

% We evaluate the efficacy of utilizing multiple different scorers to imitate different sets of relations in an environment, both when directly given, or imitating from real images. In \tbl{exp:robot_main}, we find that the use of multiple scorers
% substantially improves imitation of trajectories in the environment.

\begin{figure}[t]
\begin{center}
\includegraphics[width=1\textwidth]{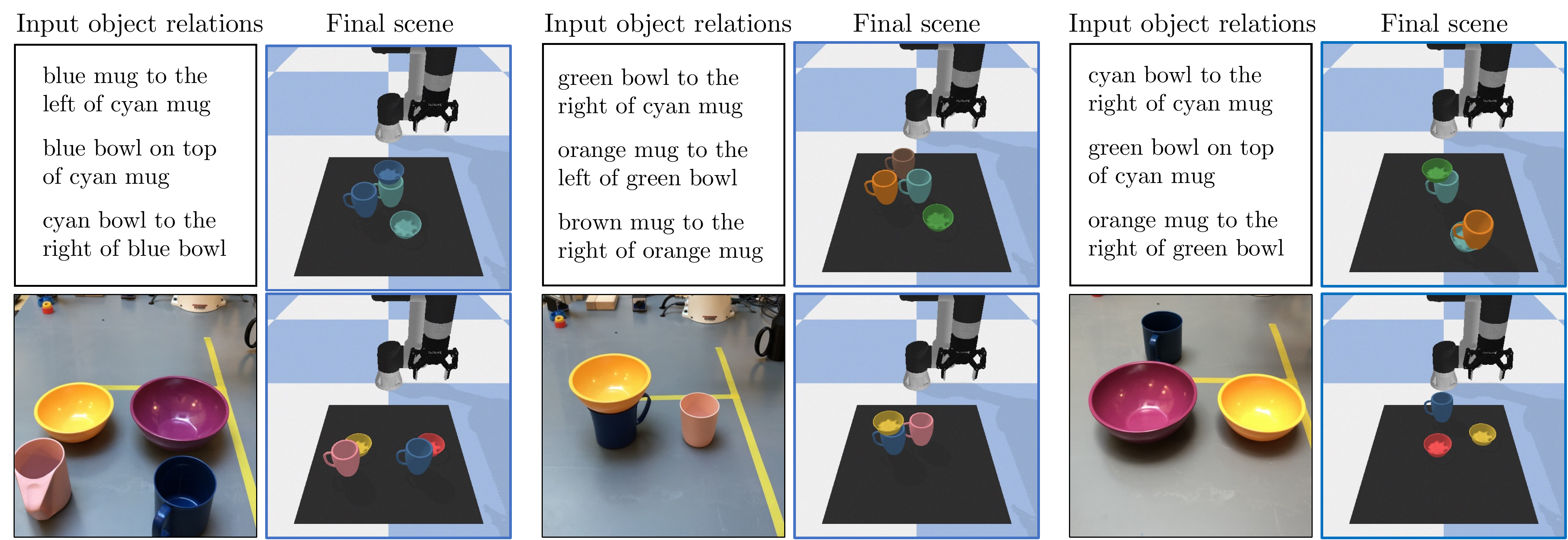}
\end{center}
\vspace{-5pt}
\caption{\small \textbf{Robot manipulation example results.} The robot manipulates objects to achieve certain object relations that are specified by textual descriptions (first row) or real-world images (second row).
}
\label{fig:robot_qual}
\vspace{-5pt}
\end{figure}

\textbf{Qualitative results.} 
\fig{fig:robot_qual} shows the example results of the proposed method manipulating objects to accomplish the given task. Our method enables zero-shot robot manipulation on objects with different sizes, colors, and shapes given either the language goal or image goal.

% Next, we illustrate qualitative results of our approach when imitating different block manipulation demonstration in the real world in \fig{fig:robot_qual}. VILD is first run on images to obtain semantic segmentations of objects in the image [CITE]. Relations are inferred from segmentation masks, and MPC is then run in the underlying environment to get a final environment state that matches the real world.

%% ---------------------------------------------------------------------

% \vspace{-3pt}
\section{Analysis}
% \vspace{-5pt}
\modelname exhibits effective zero-shot generalization ability on a variety of tasks. To further understand the source of such generalization, we investigate two key components in \modelname, \ie the composition of multiple scorers (consensus optimization) (\cref{expsec:multiple_scorers}) and the iterative refinement (\cref{expsec:iterative_refinement}).
% We also show the limitations of \modelname in \cref{expsec:limitation}.

% the effect of multiple scorers in \cref{expsec:multiple_scorers} and the effect of iterative refinement in \cref{expsec:iterative_refinement}. \cref{expsec:limitation} shows the limitation of the proposed method.

% \vspace{-3pt}
\subsection{Effect of consensus optimization}
\label{expsec:multiple_scorers}
% \vspace{-3pt}
% We further investigate the effect of composing multiple scorers on the zero-shot image generation and robot manipulation tasks.
We have shown that composing multiple scorers contributes to zero-shot generalization. We further explore the influence of gradually adding each new scorer on the zeros-shot performance.

\textbf{Image generation.} In \cref{exp:multiple_scorers_image_generation}, we first show results of composing GLIDE and the CLIP scorer. We then gradually add a new scorer, the image classifier or classifier-free guidance, each time. Finally, we report the results of composing the generator and all scorers.
The performance improves every time we add a new scorer, indicating that composing multiple scorers improves zero-shot performance.
% image generation.

\textbf{Robot manipulation.}
In \tbl{exp:robot_comp}, we analyze the effect of composing multiple scores on robot manipulation. The goal is specified by textual descriptions. Composing scores from multiple views, \modelname (G+$ \sum_{n=1}^3 E_n$) and \modelname (G+$ \sum_{n=1}^5 E_n$), leads to higher accuracy.
% $n=3$ and $n=5$

% to obtain a given set of relations in a scene -- with each scorer corresponds to a different view of the scene. We find that increasing the number of scorers correspondingly leads to increased success.

\subsection{Effect of Iterative Refinement}
\label{expsec:iterative_refinement}
% \vspace{-3pt}
Next, we explore the influence of iterative refinement on zero-shot generalization, \ie the feedback loop between the generator and scorers.
We compare \modelname with baselines that compose the generator and scorers, but with the scorers only providing feedback to the generator at the end.

% To explore the influence of iterative refinement between the generator and scorers on zero-shot tasks, 

\begin{table}[t]
\small
  \caption{\small{ \textbf{Effect of composing multiple scorers.} Image generation results on ImageNet. Gradually adding new scorers keeps improving the performance, indicating that composing multiple scorers contributes to zero-shot image generation.}}
  \vspace{-5pt}
  \label{exp:multiple_scorers_image_generation}
  \centering
  \setlength{\tabcolsep}{0.5em}
  \scalebox{0.85}{
  \begin{tabular}{lllcccccc}
    \toprule
    \bf Method Name & \bf Generator & \bf Scorer & \bf IS $\uparrow$  & \bf FID $\downarrow$ & \bf KID $\downarrow$ \\
    \midrule
    \bf \modelname (G+E1) & GLIDE & CLIP & 25.017 & 30.462 & 6.174 \\
    % GLIDE-CLS-FREE & GLIDE & CLS-FREE & 25.926 & 29.219 & 5.325 \\
    % GLIDE-CLS-GLIDE & GLIDE & CLS (GUIDED) & 22.077 & 30.871 & 7.952 \\
    % \midrule
    \bf \modelname (G+E1+E2) & GLIDE & CLIP + CLS & 30.438 & 29.543 & 5.435 \\
    \bf \modelname (G+E1+E3) & GLIDE & CLIP + CLS-FREE & 30.500 & 29.726 & 4.304 \\
    % G+E2+E3 & GLIDE & CLS-FREE + CLS (GUIDED) & 29.212 & \bf 28.151 & 5.676 \\
    % \midrule
    \bf \modelname (G+E1+E2+E3) & GLIDE & CLIP + CLS + CLS-FREE & \bf 34.952 & \bf 29.184 & \bf 3.766 \\
    \bottomrule
  \end{tabular}
  }
  \vspace{-5pt}
\end{table}

% \begin{table*}[t]
% \small
%   \caption{\small{Iterative refinement Grade school math}}
%   \vspace{-5pt}
%   \label{exp:baselines}
%   \centering
% %   \setlength{\tabcolsep}{0.5em}
%   \scalebox{0.85}{
%   \begin{tabular}{lllcccccc}
%     \toprule
%     \bf Method Name & \bf Generator & \bf Energy Scorer & \bf BS=5 \rightarrow \bf 1 (CLS) \uparrow  &  \bf BS=1 \uparrow  \\
%     \midrule
%      GPT-Pretrain & GPT-2 (Medium) (Pretrain) & - & 9.70 & 1.70 \\
%      GPT-FT & GPT-2 (Medium) (FT) & - & 14.50 & 3.50 \\
%      \midrule
%      Ours (G+E) & GPT-2 (Medium) (Pretrain) & CLS & \bf 17.20 & \bf 16.80 \\
%     \bottomrule
%   \end{tabular}
%   }
%   \vspace{-5pt}
% \end{table*}

% \begin{table*}[t]
% \small
%   \caption{\small{Iterative refinement Grade school math}}
%   \vspace{-5pt}
%   \label{exp:iterative_refinment}
%   \centering
% %   \setlength{\tabcolsep}{0.5em}
%   \scalebox{0.85}{
%   \begin{tabular}{lllcccccc}
%     \toprule
%     \bf Method Name & \bf Generator & \bf Energy Scorer &  \bf BS=1 \uparrow & \bf BS=5 \rightarrow \bf 1 (CLS)  \\
%     \midrule
%      GPT-Pretrain & GPT-2 (Medium) (Pretrain) & - & 1.744 & 9.704 \\
%      GPT-FT & GPT-2 (Medium) (FT) & - & 3.487 & 14.481 \\
%      \midrule
%      Ours (G+E) & GPT-2 (Medium) (Pretrain) & CLS & \bf 16.831 & \bf 17.210 \\
%     \bottomrule
%   \end{tabular}
%   }
%   \vspace{-5pt}
% \end{table*}

\begin{table*}[t]
\small
  \caption{\small{ \textbf{Effect of iterative refinement.} 
  Grade school math results on GSM8K. \modelname with iterative refinement outperforms baselines where the scorer only provides feedback to the generator at the end stage ($t=T$). BS is the beam search size.
%   Iterative communication between the generator and scorers enables the models to correct errors caused by other models, outperforming baselines without iterative refinement. 
  }}
  \vspace{-5pt}
  \label{exptab:iterative_refinment}
  \centering
  \scalebox{0.85}{
  \begin{tabular}{lllcccccc}
    \toprule
    \bf Method Name & \bf Generator & \bf Scorer & \bf Interaction & \bf BS=1 $\uparrow$  \\
    \midrule
     \bf GPT-Pretrain+E & GPT-2 (Medium) (Pretrain) & CLS & $t=T$ & 9.704 \\
     \bf GPT-FT+E & GPT-2 (Medium) (FT) & CLS & $t=T$ & 14.481 \\
     \midrule
     \bf \modelname (G+E) & GPT-2 (Medium) (Pretrain) & CLS & $t=\{1,\cdots, T\}$ & \bf 17.210 \\
    \bottomrule
  \end{tabular}
  }
  \vspace{-5pt}
\end{table*}

% \begin{table*}[t]
% \small
%   \caption{\small{Robot performance when imitating different trajectories}}
%   \vspace{-5pt}
%   \label{exp:robot_comp}
%   \centering
%   \setlength{\tabcolsep}{0.5em}
%   \scalebox{0.85}{
%   \begin{tabular}{l|c|cccc}
%     \toprule
%     \bf \multirow{2}{*}{Model} & \bf \multirow{2}{*}{Optimization Steps}   &   \multicolumn{2}{c}{\bf 2 Relations} &  \multicolumn{2}{c}{\bf 3 Relations}\\
%      &    &  Relation $\uparrow$ & Overall $\uparrow$  &   Relation $\uparrow$ & Overall $\uparrow$ \\
%     \midrule
%      \bf \modelname (G+E1)  & 25  & 50.0 & 35.0 & 75.0 & 50.0   \\
%     \bf \modelname (G+E1+E2+E3) & 25  & 65.0 & 57.5 & 78.3 &  63.3 \\
%     \bf \modelname (G+E1+E2+E3+E4+E5) & 25 & 72.5 & 67.5 & 85.0 & 75.0 \\
%     \midrule
%       \bf \modelname (G+E1+E2+E3) & 10  & 57.5  & 52.5  & 68.3 & 53.3 \\
%       \bf \modelname (G+E1+E2+E3) & 25 & 65.0  &  57.5 & 78.3 & 63.3 \\
%       \bf \modelname (G+E1+E2+E3) & 50  & 85.0 &  77.5 & 85.0 & 75.0 \\
%     \bottomrule
%   \end{tabular}
%   }
%   \vspace{-5pt}
% \end{table*}

\begin{wraptable}{r}{8cm}
\small
\renewcommand{\arraystretch}{1.2}
  \vspace{-7pt}
  \caption{\small{\textbf{Effect of composing multiple scorers and iterative refinement on robot manipulation.} Both components are important for zero-shot generalization.}}
  \vspace{-7pt}
  \label{exp:robot_comp}
  \centering
  \setlength{\tabcolsep}{0.5em}
  \scalebox{0.85}{
  \begin{tabular}{l|c|cccc}
    \toprule
    \bf Method Name & \bf Interaction  &   \bf 2 Relations &  \bf 3 Relations \\
    \midrule
     \bf \modelname (G+E1)  & $t=\{1,\cdots, T\}$ & 35.0 & 50.0   \\
    \bf \modelname (G+$\bf \sum_{n=1}^3 E_n$) & $t=\{1,\cdots, T\}$  & 57.5 &  63.3 \\
    \bf \modelname (G+$\bf \sum_{n=1}^5 E_n$) & $t=\{1,\cdots, T\}$ & \bf 67.5 & \bf 75.0 \\
    \midrule
    \bf No-IR (G+$\bf \sum_{n=1}^5 E_n$) & $t=T$  & 30.0 & 46.6\\
    \bottomrule
  \end{tabular}
  }
  \vspace{-5pt}
\end{wraptable}

\textbf{Grade school math.}
% We compared the proposed method with two baselines, \emph{GPT-Pretrain} and \emph{GPT-FT}, in \cref{expsec:grade_school_math}. These two baselines generate math solutions without the feedback from scorers.
% In this experiment, we add scorers to the baselines to allows them provide feedback to the generator. Different from our method, the interaction between the generator and scorers of these baselines only happens in the end stage.
%
% We compared the proposed method (using iterative refinment) with baselines that compose the generator and scorer, but only refine the generate results in the end stage. 
In \cref{exptab:iterative_refinment}, the baselines, GPT-Pretrain+E and GPT-FT+E, generate five proposal solutions of a given math problem. Then the scorer, \ie the same question-solution classifier used in \modelname, selects the best solution based on its score.
\modelname iteratively refines the generated answer while the baselines refine the entirely generated solutions in the end.
% (\ie refines each word in the answer)
\modelname and GPT-Pretrain+E use the same generator and scorer, but \modelname outperforms GPT-Pretrain+E by $\%7.507$.
\modelname still achieves better performance than GPT-FT+E, which uses a stronger generator (finetuned on the GSM8K dataset).

\textbf{Robot manipulation.} In \tbl{exp:robot_comp}, the baseline, No-IR (G+$\sum_{n=1}^5 E_n$), first samples 100 trajectories without using the feedback from scorers. Then the scorers select the best trajectories based on the summed score.  
The generator and scorers of this baseline are the same as our method, \ie \modelname (G+$\sum_{n=1}^5 E_n$), but our method outperforms the baseline by $\%37.5$ on the ``2 Relations'' setting, indicating the effectiveness of iterative refinement in the proposed framework. 
% we analyze the effect of utilize different number of steps to optimize a given set of relations in a scene.  As more steps of optimization are run, the corresponding performance of our approach improves.

Together, these results show that the composition of multiple scorers and iterative refinement are both important for zero-shot generalization.
These results point to the potential broader applicability of the proposed method as a general purpose framework for zero-shot multimodal tasks.

% \vspace{-3pt}
\section{Conclusion and Future Work}
% \vspace{-3pt}

In this paper, we propose a unified framework for composing ensembles of pre-trained models through iterative consensus without any training or finetuning. Our framework consists of a generator and an ensemble of scorers. The scorers provide feedback to the generator to iteratively improve its generated results.
We show the proposed method allows effective zero-shot generalization on four representative tasks, \ie image generation, video question answering, grade school math, and robot manipulation, and even outperforms methods that directly finetune models on certain tasks. 
We further analyze the source of such zero-shot generalization by exploring the effect of the composition of multiple scorers and the iterative refinement, and find that both are important for zero-shot generalization.

As our method does not need any training or finetuning, one drawback is that its performance depends on the pre-trained models. 
% Even though the diverse composition of generators and energy scorers can be used for solving many new tasks, our method cannot fundamentally change the property of the generators. 
% For example, GLIDE cannot generate scenes containing certain object relations, such as ``a large green sphere on top of a small blue cube''. Composing a pre-trained image-object classifier (scorer) cannot make the generator generate such scenes either. 
% Thus effective zero-shot generalization also requires powerful pre-trained models.
Training large models are complementary to the framework and methods we proposed and may be directly applied. We hope to explore these directions for zero-shot generalization in future work.
In addition, our framework enables the composition of separately trained models and boosts performance by leveraging the knowledge from multiple expert models.
The scorers can be learned at different times on different data in an incremental-learning manner, enabling the combination of incrementally learned knowledge.
Our framework thus paves the way for many potential applications in lifelong learning / continual learning settings.
% Our framework naturally supports such lifelong learning / continual learning settings, which is interesting to explore in future work.

\textbf{Acknowledgments.} Shuang Li is partially supported by Meta Research Fellowship.
This research is partially supported by the US Army, under the DEVCOM Army Research Laboratory project, reg. no. 1130233-442111. The content does not necessarily reflect the position or the policy of any government,
and no official endorsement should be inferred. Yilun Du is supported by a NSF Graduate Fellowship.

\clearpage

\bibliography{iclr2023_conference}
\bibliographystyle{iclr2023_conference}

\clearpage

\renewcommand{\thesection}{A.\arabic{section}}
\renewcommand{\thefigure}{A\arabic{figure}}
\renewcommand{\thetable}{A\arabic{table}}

\setcounter{section}{0}
\setcounter{figure}{0}
\setcounter{table}{0}

\renewcommand{\theequation}{A\arabic{equation}}
\setcounter{equation}{0}

\newpage
\appendix
\textbf{\Large{Appendix}}
\vspace{5pt}

% \appendix
% \section{Appendix}

In this appendix, we first show experimental details of each task in \cref{apx:task_details}. We then show the ethics statement of the Amazon Mechanical Turk experiment for video question answering in \cref{apx:mturk_exp}.

\section{Experimental details}
\label{apx:task_details}

In this section, we provide more experimental details of each task.
We use TITAN RTX 24GB GPUs for all the experiments. 

\subsection{Image generation}
\label{apx_sec:image_generation}
We use the reverse diffusion process of GLIDE, a text-guided diffusion model, as the generator to generate image proposals. 
At each step of the diffusion process (corresponding to a step of the iterative refinement), we use the gradient from an ensemble of scorers to guide and update the generated proposals. We iteratively repeat this procedure until the final step.

\begin{figure}[t]
\begin{center}
\includegraphics[width=0.5\textwidth]{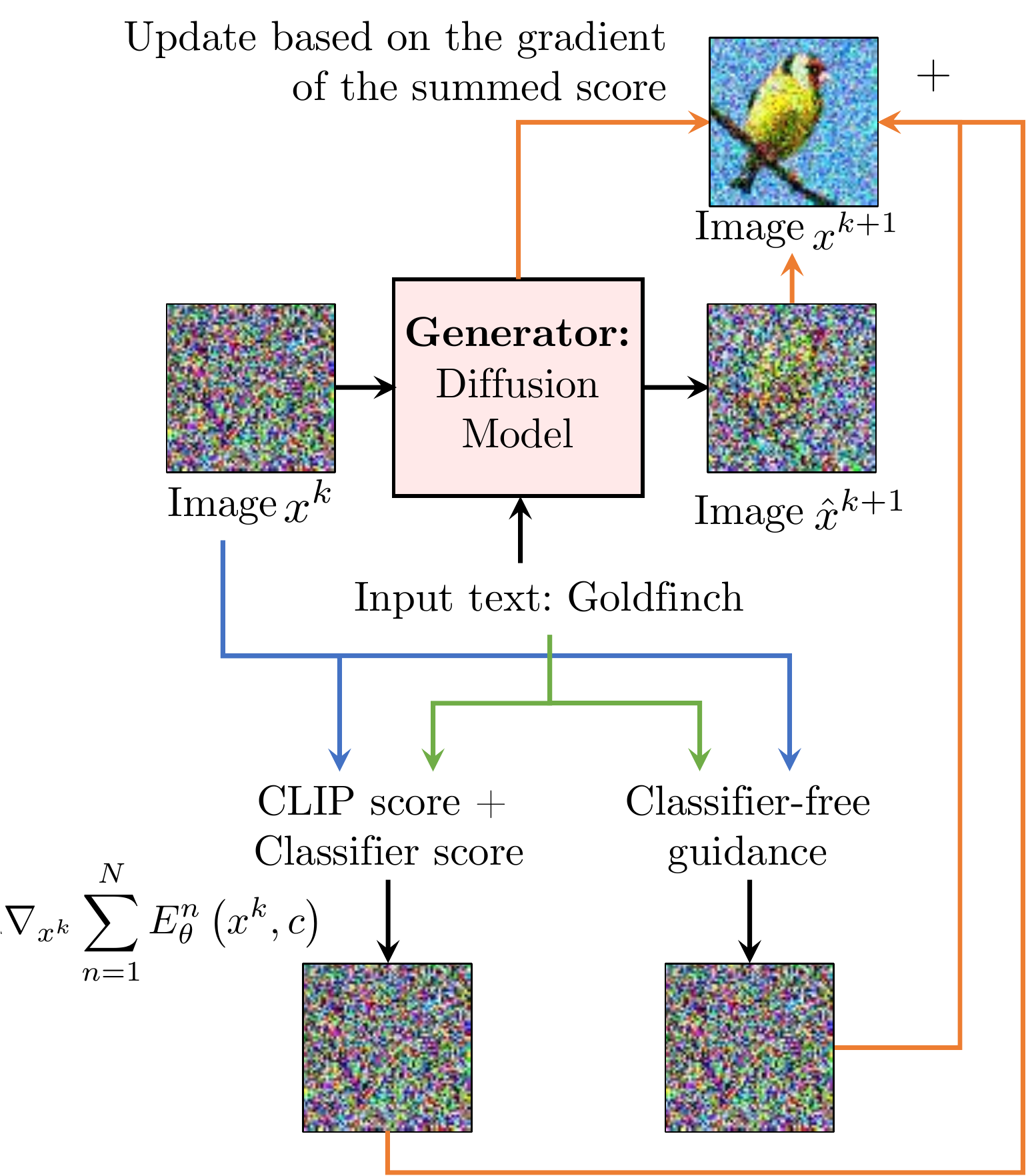}
\end{center}
\vspace{-5pt}
\caption{\small \textbf{Overview of image generation.} We use the reverse diffusion process of GLIDE~\citep{nichol2021glide}, a text-guided diffusion model, as the generator to generate image proposals. 
At each step of the diffusion process (corresponding to a step of the iterative refinement), we use the gradient from an ensemble of scorers, such as CLIP~\citep{radford2021learning}, to guide and update the generated proposals. 
The image $x^{k}$ generated at iteration $k$ is first sent to the diffusion model to generate an image proposal $\hat{x}^{k+1}$. 
The scorers provide feedback to refine the generated result. 
The CLIP model computes the cosine similarity between the image and text features as the score. The image classifier~\citep{dhariwal2021diffusion} predicts the probability of the image matching the text label as the score. The scores generated by different scorers are summed, and their gradient with respect to $x^{k}$ is used to compute the next reverse prediction $x^{k+1}$.
Classifier-free guidance ~\citep{ho2022classifier} can be treated as an implicit classifier that directly provides pixel-wise gradient feedback to the generated image.
We iteratively repeat this procedure until the final step.
Our framework enables the use of ensembles of different pre-trained models as scorers, significantly improving the zero-shot results by leveraging the strengths of multiple expert models.
}
\label{apxfig:image_generation}
% \vspace{-5pt}
\end{figure}

As shown in \cref{apxfig:image_generation}, the image $x^{k}$ generated at iteration $k$ is first sent to the diffusion model to generate an image proposal $\hat{x}^{k+1}$. 
The scorers provide feedback to refine the generated result. 
The CLIP model computes the cosine similarity between the image and text features as the score (we used the pre-trained CLIP model from \citep{ho2022classifier}.). The image classifier~\citep{dhariwal2021diffusion} predicts the probability of the image matching the text label as the score. The scores generated by different scorers are summed, and their gradient with respect to $x^{k}$ is used to compute the next reverse prediction $x^{k+1}$.
The classifier-free guidance~\citep{ho2022classifier} can be treated as an implicit classifier that directly provides pixel-wise gradient feedback to the generated image.
Our framework enables the use of ensembles of different pre-trained models as scorers, significantly improving the zero-shot results by leveraging the strengths of multiple expert models.

Our implementation for image generation is modified based on the code of GLIDE~\citep{nichol2021glide} and the classifier guidance diffusion~\citep{dhariwal2021diffusion}. 
We use DDIM to sample images from GLIDE in 100 steps. The guidance scale is set to 3. 
% The diffusion steps are 100.

\subsection{Video question answering}
\label{apx_sec:vqa}
In video question answering, we use the proposed method to generate captions for the video frames and then use GPT-3 to summarize the captions to answer questions. 
We use GPT-2 as the generator and a set of CLIP models as scorers to generate captions for each video frame.
The CLIP models~\citep{radford2021learning,reimers-2019-sentence-bert} are from the Huggingface library~\citep{wolf2019huggingface}:
\begin{itemize}[leftmargin=1.2em]
    \item CLIP-32: \href{https://huggingface.co/openai/clip-vit-base-patch32}
    {\texttt{https://huggingface.co/openai/clip-vit-base-patch32}}.
    \item CLIP-14: \href{https://huggingface.co/openai/clip-vit-large-patch14}
    {\texttt{https://huggingface.co/openai/clip-vit-large-patch14}}.
    \item CLIP-multilingual: \href{https://huggingface.co/sentence-transformers/clip-ViT-B-32-multilingual-v1}
    {\texttt{https://huggingface.co/sentence-transformers/clip-ViT-B-32-multilingual-v1}}.
\end{itemize}

\begin{figure}[t]
\begin{center}
\includegraphics[width=0.78\textwidth]{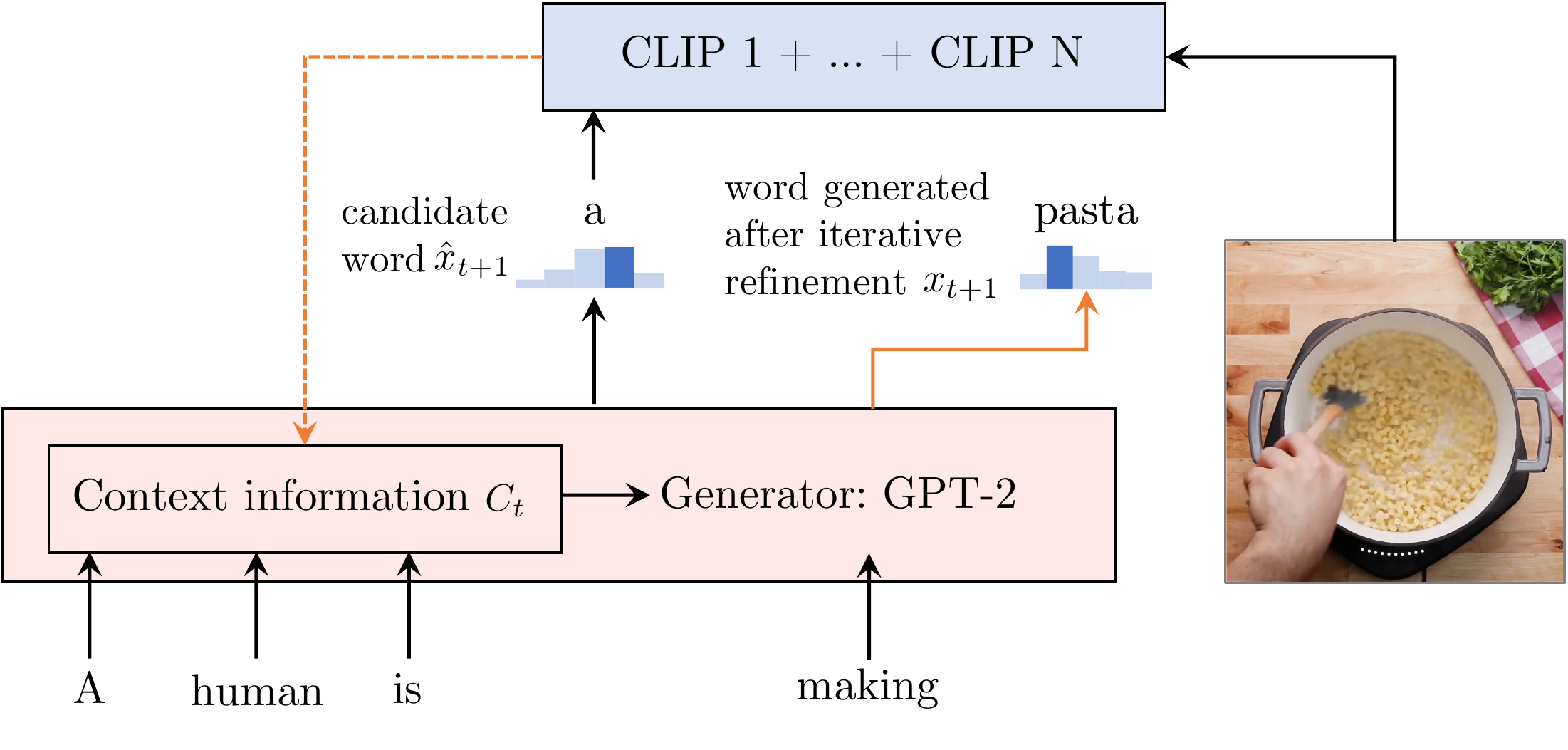}
\end{center}
\vspace{-15pt}
\caption{\small \textbf{Overview of video frame captioning for video question answering.} 
We use GPT-2 as the generator and a set of CLIP models as scorers to generate captions for each video frame.
% Given a video frame $I$, we generate a sequence of words to describe it. 
To integrate feedback from scorers to the generator, similar to ZeroCap~\citep{tewel2021zero}, we define a context cache $C_t$ (a set of embedding functions in GPT-2) that stores the context information generated so far, which is updated iteratively based on the feedback from scorers. 
To update $C_t$, we first use $G$ to generate a set of candidate words $\hat{X}_{t+1}=\{\hat{x}_{t+1}\}$, and then use the feature distance (after softmax) between each sentence (the concatenation of previous words and each new word $\{x_1, x_2, \cdots, \hat{x}_{t+1}\}$, where $\hat{x}_{t+1} \in \hat{X}_{t+1}$) and the video frame as the probability of them matching. The CLIP score is the cross-entropy loss $\mathcal{L_{\text{CLIP}}}$ between this new probability distribution and the original distribution of the next word obtained from the generator $G$ (see Equation 4 in \citep{tewel2021zero}).
The gradient of summed scores (multiple CLIP models) is propagated to $G$ to update $C_t$ (see Equation 5 in \citep{tewel2021zero}).
After several iterations, the updated $C_t$ is used to generate the next token $x_{t+1}=G(x_t, C_t)$. We repeat this process until we generate the entire caption.
We cascade the captions of multiple video frames and questions about this video to prompt GPT-3 for video question answering.
}
\label{apxfig:vqa}
\vspace{-5pt}
\end{figure}

\cref{apxfig:vqa} shows the framework for generating frame captions.
Given a video frame $I$, we generate a sequence of words to describe it. To integrate feedback from scorers to the generator, similar to ZeroCap~\citep{tewel2021zero}, we define a context cache $C_t$ (a set of embedding functions in GPT-2) that stores the context information generated so far, which is updated iteratively based on the feedback from scorers. 
The prediction of the next word from the generator $G$ is given by $x_{t+1}=G(x_t, C_t)$. 
% To update $C_t$, we first use $G$ to generate a set of candidate words $\hat{X}_{t+1} = \{\hat{x}_{t+1}\}$, and then use the feature distance (after softmax) between each sentence (the concatenation of previous words and each new word $\{x_1, x_2, \cdots, \hat{x}_{t+1}\}$, where $\hat{x}_{t+1} \in \hat{X}_{t+1}$) and the video frame as the probability of them matching. The CLIP score is the cross-entropy loss $\mathcal{L_{\text{CLIP}}}$ between this new probability distribution and the original distribution of the next word obtained from the generator $G$.

Our implementation is based on the code of ZeroCap~\citep{tewel2021zero}. 
The context cache $C_t$ is updated in the same way as Equation 5 in \citep{tewel2021zero}, but we compose multiple CLIP scores when providing the feedback to $C_t$.
The CLIP loss $\mathcal{L}_{\text{CLIP}}$ is similar to their Equation 4.
We also used the cross-entropy loss $\mathcal{L}_{\text{CE}}$ in their Equation 2 to ensure the generated sentence is grammatically sound. 
After several iterations, the updated $C_t$ is used to generate the next token $x_{t+1}=G(x_t, C_t)$. We repeat this process until we generate the entire caption.

\begin{figure}[t]
\begin{center}
\includegraphics[width=0.85\textwidth]{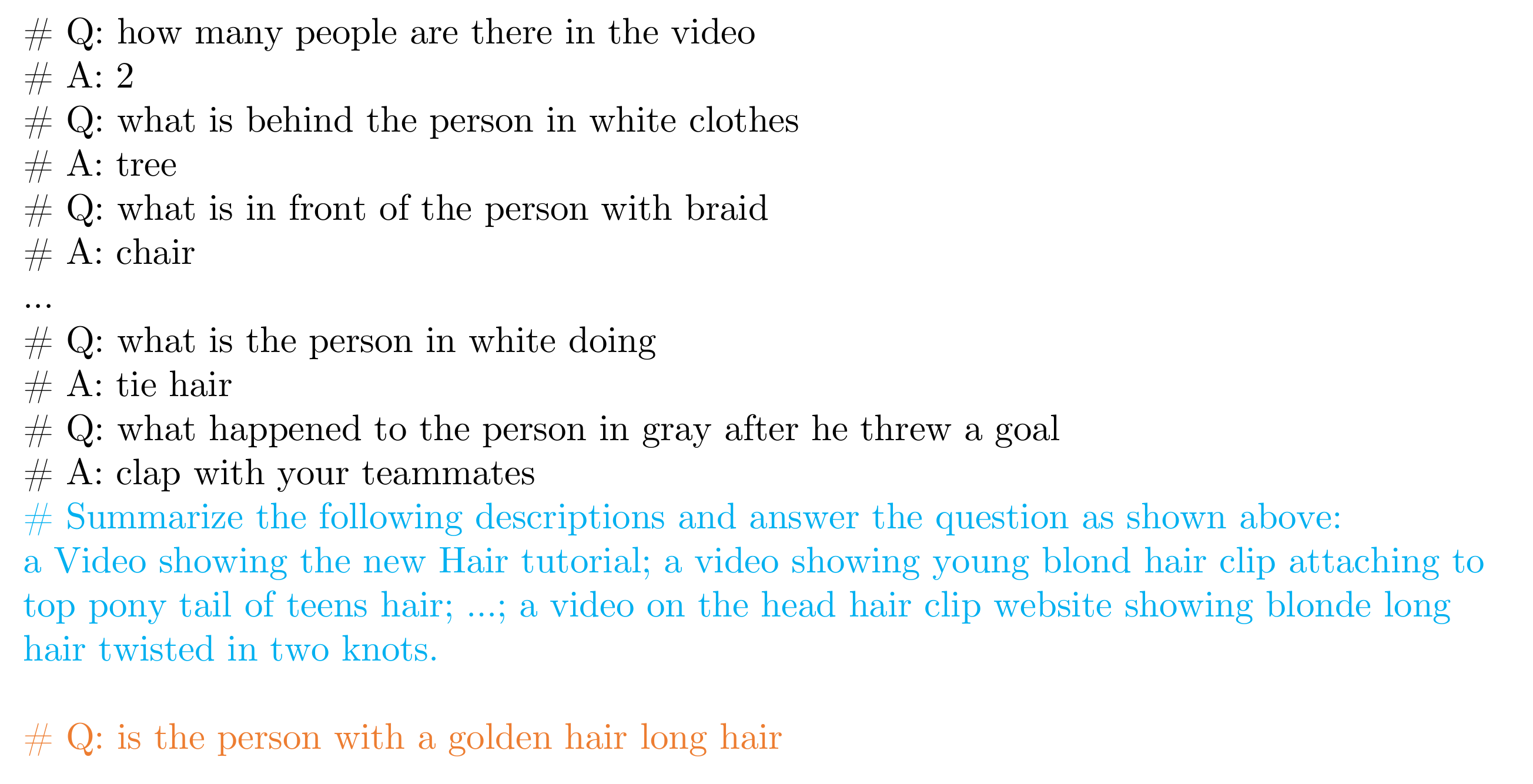}
\end{center}
\vspace{-15pt}
\caption{\small \textbf{Prompt given to GPT-3 for video question answering.} Text in black contains the question-answer pairs randomly sampled from the ActivityNet-QA training dataset. Text in blue has the video frame captions generated by the proposed method. Text in orange is the question about this video that needs to be answered.
}
\label{apxfig:vqa_qa}
\vspace{-5pt}
\end{figure}

To answer the video questions, we cascade the generated captions of the video frames and the questions about this video to prompt GPT-3 to generate answers.
For each video, we delete the first 10 frames and the last 10 frames to remove the beginning or ending advertisements. We then take 30 video frames evenly from the rest frames and send them to GPT-3.
To guide GPT-3 to generate proper answers, we randomly select 30 question-answer pairs from the training set of ActivityNet-QA~\citep{yu2019activitynet} and use them as part of the prompt of GPT-3.
As shown in \cref{apxfig:vqa_qa}, the prompt of GPT-3 consists of examples of question-answer pairs, the video frame captions generated by the proposed method, and the question about this video that needs to be answered. The text generated by GPT-3 is used as the answer to the question asked. We also used the profanity check tool (\href{https://github.com/vzhou842/profanity-check}{\texttt{https://github.com/vzhou842/profanity-check}}) to remove the improper answers.

% The goal is to update $C_t$ iteratively based on the CLIP score to generate the next word such that the sentence is grammatically sound as well as accurately describes the given video frame. 

% , but do not show it for simplicity. 

% We use a similar approach as \citep{tewel2021zero} to generate zero-shot image captions, but our focus of this paper is to propose a general framework for composing pre-trained models across a variety of tasks.
% Similar to \citep{tewel2021zero}, we use GPT-2 and CLIP to generate image captions. 
% However, our focus of this paper is to propose a general framework for composing pre-trained models across a variety of tasks, and leverage the knowledge from multiple expert models to improve the performance.

\subsection{Grade school math}
\begin{figure}[t]
\begin{center}
\includegraphics[width=0.8\textwidth]{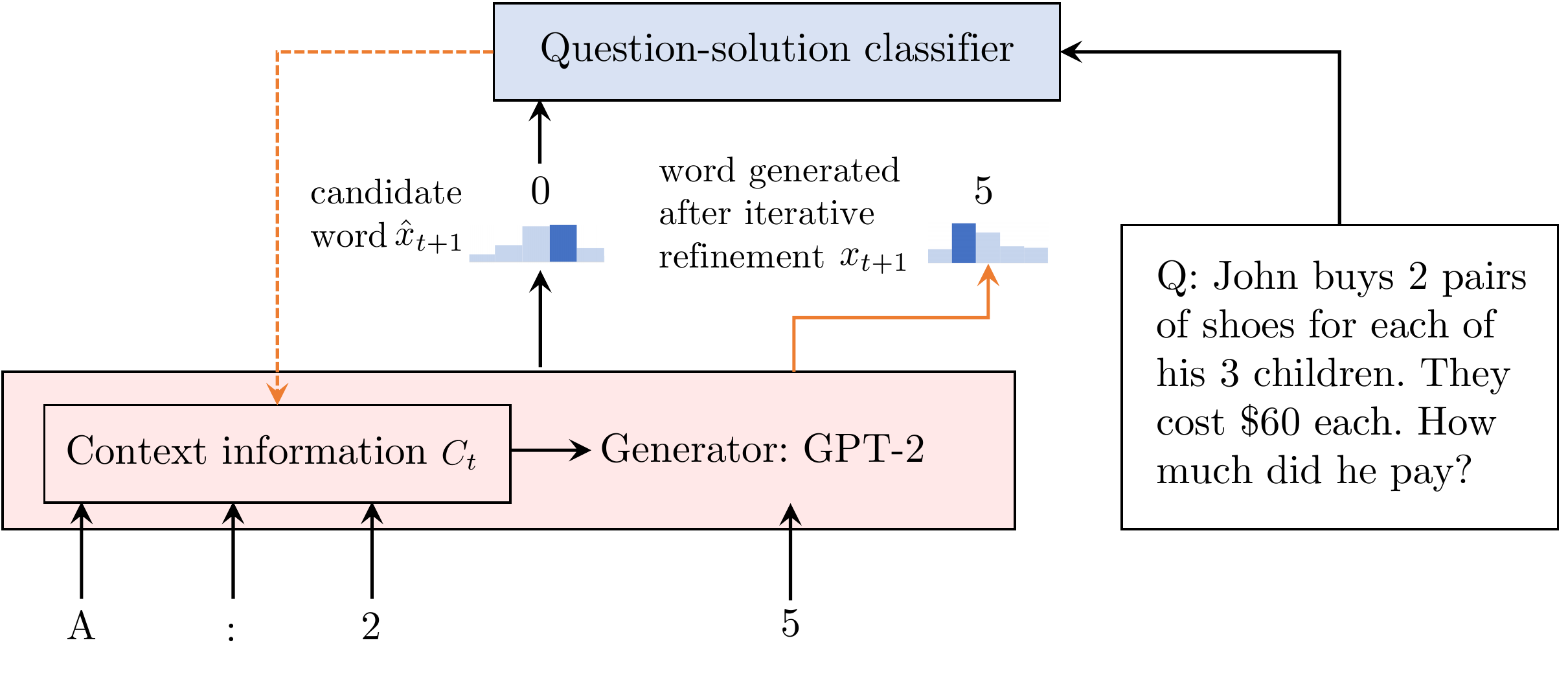}
\end{center}
\vspace{-10pt}
\caption{\small \textbf{Overview of solving grade school math problems.} We use GPT-2 as the generator and treat the grade school math problem as a text generation problem. The scorer, a pre-trained question-solution classifier, provides the generator feedback to guide the next token's generation $x_{t+1}$. We follow the approach used in VQA to iteratively optimize the generations based on the feedback from scorers. Our generator $G$ first generates a set of candidate words $\hat{X}_{t+1}=\{\hat{x}_{t+1}\}$, and then the classifier predicts the probability of each solution (the concatenation of previous words and each new word $\{x_1, x_2, \cdots, \hat{x}_{t+1}\}$, where $\hat{x}_{t+1} \in \hat{X}_{t+1}$) matching the given question. 
The classifier score is the cross-entropy loss between this new probability distribution and the original distribution of the next word obtained from the generator $G$. The gradient of the classifier score is used to update $C_t$ through iterative refinement (see Equation 5 in \citep{tewel2021zero}). The updated $C_t$ is used to predict the next word $x_{t+1}=G(x_t, C_t)$. We repeat this process until we generate the complete solution.
}
\label{apxfig:math}
% \vspace{-5pt}
\end{figure}

We treat the grade school math problem as a text generation problem. As shown in \cref{apxfig:math}, we use GPT-2 as the generator and a pre-trained question-solution classifier as the scorer.
The pre-trained classifier is a binary classifier trained on the training set of GSM8K~\citep{cobbe2021training}. 
Given a math problem, such as ``Natalia sold clips to 48 of her friends in April, and then she sold half as many clips in May. How many clips did Natalia sell altogether in April and May?'', and an answer, such as ``72''. If the answer is correct for the given problem, then the label is 1; otherwise, the label is 0. 

After training, the classifier is used as the scorer to provide feedback to the generator to guide the next token's generation $x_{t+1}$. Similar to VQA, the generator $G$ first generates a set of candidate words $\hat{X}_{t+1} = \{\hat{x}_{t+1}\}$, and then the classifier predicts the probability of each solution (the concatenation of previous words and each new word $\{x_1, x_2, \cdots, \hat{x}_{t+1}\}$, where $\hat{x}_{t+1} \in \hat{X}_{t+1}$) matching the given question. 
The classifier score is the cross-entropy loss between this new probability distribution and the original distribution of the next word obtained from the generator $G$ (the way to compute the classifier score is the same as computing the CLIP score in VQA). 
We also used the cross-entropy loss $\mathcal{L}_{\text{CE}}$ in Equation 2 of ZeroCap~\citep{tewel2021zero} to ensure the generated sentence is grammatically sound. 
The context cache $C_t$ is updated in the same way as Equation 5 in \citep{tewel2021zero}, but we use the classifier score when providing the feedback to $C_t$.
The updated $C_t$ is used to predict the next word $x_{t+1}=G(x_t, C_t)$. We repeat this process until we generate the complete solution.

\begin{figure}[t]
\begin{center}
\includegraphics[width=1\textwidth]{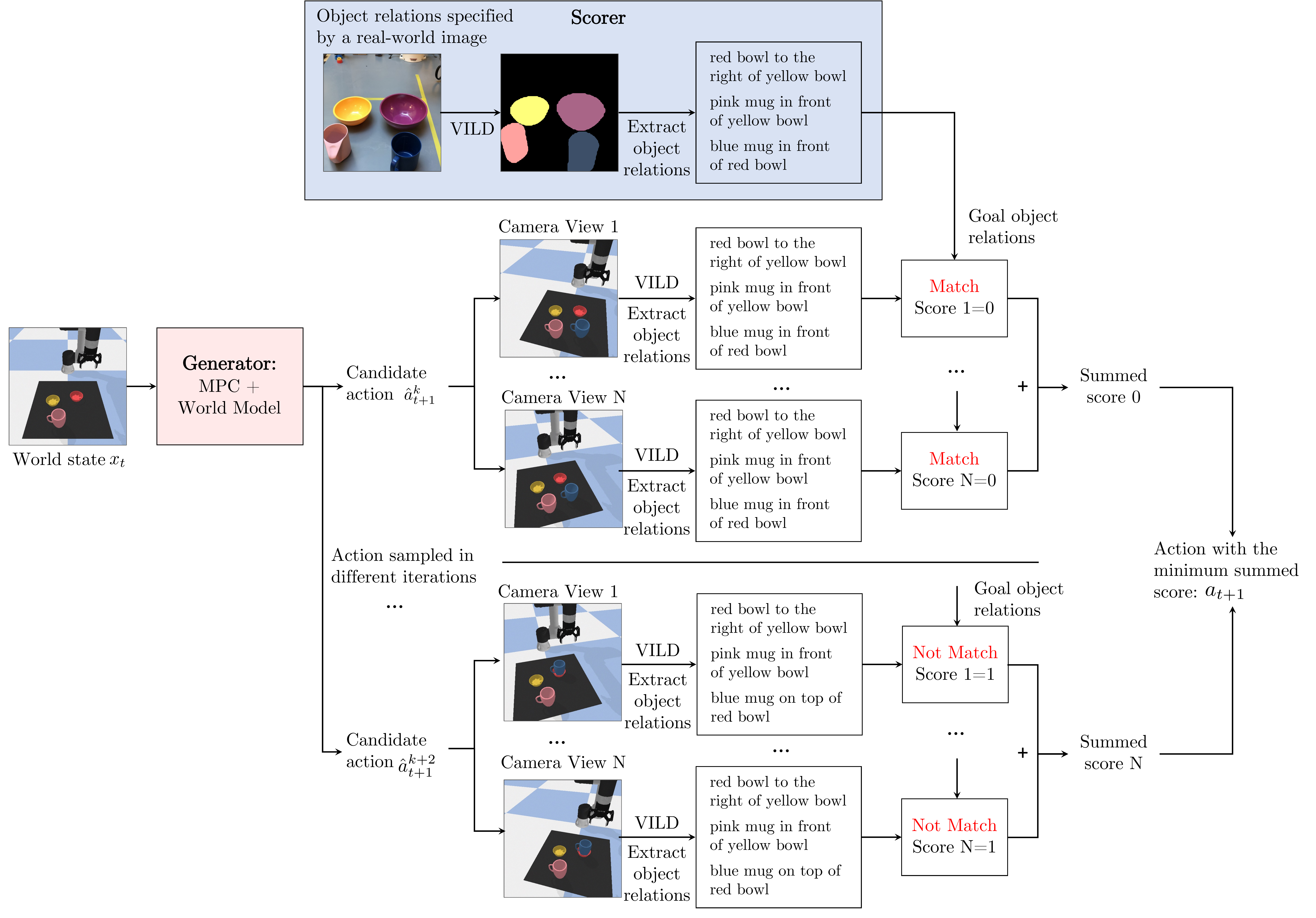}
\end{center}
% \vspace{-10pt}
\caption{\small \textbf{Overview of robot manipulation.}
We use MPC+World Model as the generator and ViLD as the scorer to manipulate objects to conform to a set of object relations specified by text descriptions or real-world images. 
\textbf{Top:} given a real-world image, we first use ViLD to generate a 2D segmentation of the real-world image and the corresponding text label, such as ``mug'', for each segment. We then use the relative pixel-wise offsets of segmentation masks and the text labels to infer a set of object relations.
\textbf{Bottom:} Given the current world state $x_t$, we aim to generate an action $a_{t+1}$ so that the new world state after executing $a_{t+1}$ has object relations closer to the object relations in the given image.
To do this, we first use the generator (MPC+World model) to generate a set of candidate actions $\{\hat{a}_{t+1}^k\}$ and the corresponding world states $\{\hat{x}_{t+1}^k\}$ after executing each candidate action.
For each new world state $\hat{x}_{t+1}^k$, we render $N$ 2D images from $N$ camera views. Each rendered image is sent to VILD to get a segmentation map and text labels. We project the objects into 3D space based on the segmentation map and the depth map of the image. We then obtain the object relations based on their 3D positions and predicted text labels. We compare the object relations obtained from each rendered image and the object relations obtained from the real-world image to compute the score. The score is 0 if the relations are matching; otherwise, 1. We sum the scores from each rendered image to obtain the final score.
We choose the action $a_{t+1}$ that leads to a world state with the minimum summed score. We execute $a_{t+1}$ in the environment and get a new state $x_{t+1}$. We repeat this process until the task is accomplished or we are at the final step $T$.
% When executing relations from a real world image -- a set of segmentation of objects are extracted using VILD. Relations are inferred from pixel-wise segmentations from VILD, and are used as target relations for MPC. 
}
\label{apxfig:robot}
% \vspace{-5pt}
\end{figure}

\subsection{Robot manipulation}
\label{apx_sec:robot}

In robot manipulation, we use the proposed method to manipulate objects in Ravens~\citep{zeng2020transporter} to conform to a set of object relations specified by text descriptions or real-world images. 
We use MPC+World Model as the generator and ViLD~\citep{gu2021open} as the scorer.
As shown in \fig{apxfig:robot}, given a real-world image, our model manipulates objects in the environment to achieve a state with objects having the same object relations as the given image.
We first use ViLD to generate a 2D segmentation of the real-world image and the corresponding text label, such as ``mug'', for each segment. We then use the relative pixel-wise offsets of segmentation masks and the text labels to infer a set of object relations (top panel of \fig{apxfig:robot}). 

Given the current world state $x_t$, we aim to generate an action $a_{t+1}$ so that the new world state after executing $a_{t+1}$ has object relations closer to the object relations in the given image.
To do this, we first use the generator (MPC+World Model) to generate a set of candidate actions $\{\hat{a}_{t+1}^k\}$ and the corresponding world states $\{\hat{x}_{t+1}^k\}$ after executing each candidate action. For each new world state $\hat{x}_{t+1}^k$, we render $N$ 2D images from $N$ camera views. Each rendered image is sent to VILD to get a segmentation map and text labels. We project the objects into 3D space based on the segmentation map and the depth map of the image. We then obtain the object relations based on their 3D positions and the predicted text labels. We compare the object relations obtained from each rendered image and the object relations obtained from the real-world image to compute the score. The score is 0 if the relations are matching; otherwise, 1. We sum the scores from each rendered image to obtain the final score.
We choose the action $a_{t+1}$ that leads to a world state with the minimum summed score. We execute $a_{t+1}$ in the environment and get a new state $x_{t+1}$. We repeat this process until the task is accomplished or we are at the final step $T$, where $T$ equals to the number of relations extracted from the real-world image.

\subsection{A unified framework for composing pre-trained models}
Our method shares some similar architecture with existing works, such as ZeroCap~\citep{tewel2021zero} and CLIP-guided diffusion models~\citep{nichol2021glide}. However, the focus of our paper is to propose a general framework for composing different pre-trained models across a variety of tasks, and these particular methods are concrete instantiations of our proposed framework. In addition, in this work, we also illustrate how we may combine ensembles of different pre-trained models as scorers to leverage the ``wisdom of the crowds'' where each scorer provides complementary feedback to the generator, compensating for the potential weaknesses of other scorers. 
% which may provide complementary feedback to the generator and compensate for the weaknesses of an individual model. 
Through iterative optimization and the composition of multiple scorers, our method shows effective zero-shot generalization ability on various multimodal tasks.

\clearpage

\section{Ethics Statement of Amazon Mechanical Turk Experiments}
\label{apx:mturk_exp}
\begin{figure}[t]
\begin{center}
\includegraphics[width=1\textwidth]{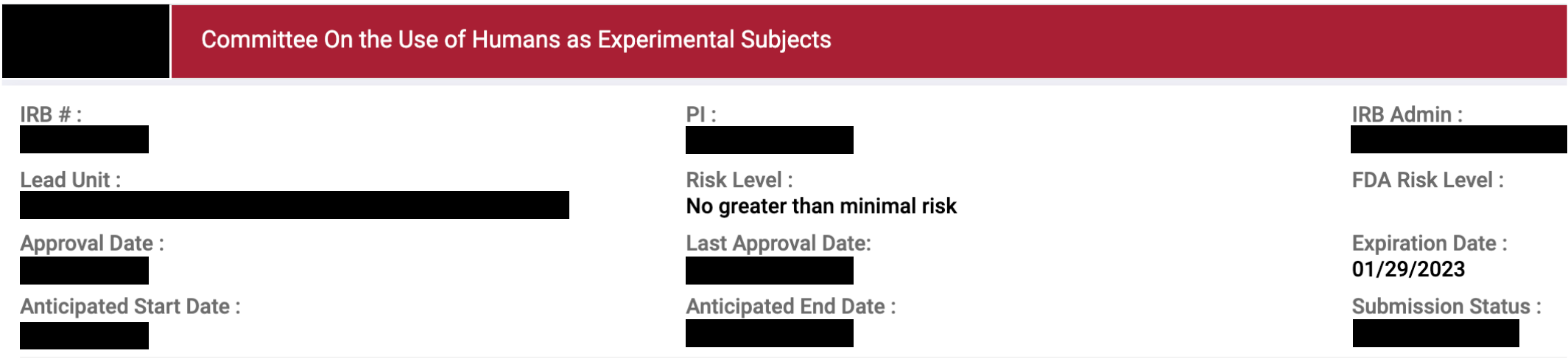}
\end{center}
\vspace{-5pt}
\caption{\small Screenshot of the approval form from the Committee on the Use of Humans as Experimental Subjects.
% If the answer matches the given video and question, the worker should choose ``yes''; otherwise, ``no''.
}
\label{apxfig:irb}
% \vspace{-5pt}
\end{figure}

\begin{figure}[t]
\begin{center}
\includegraphics[width=1\textwidth]{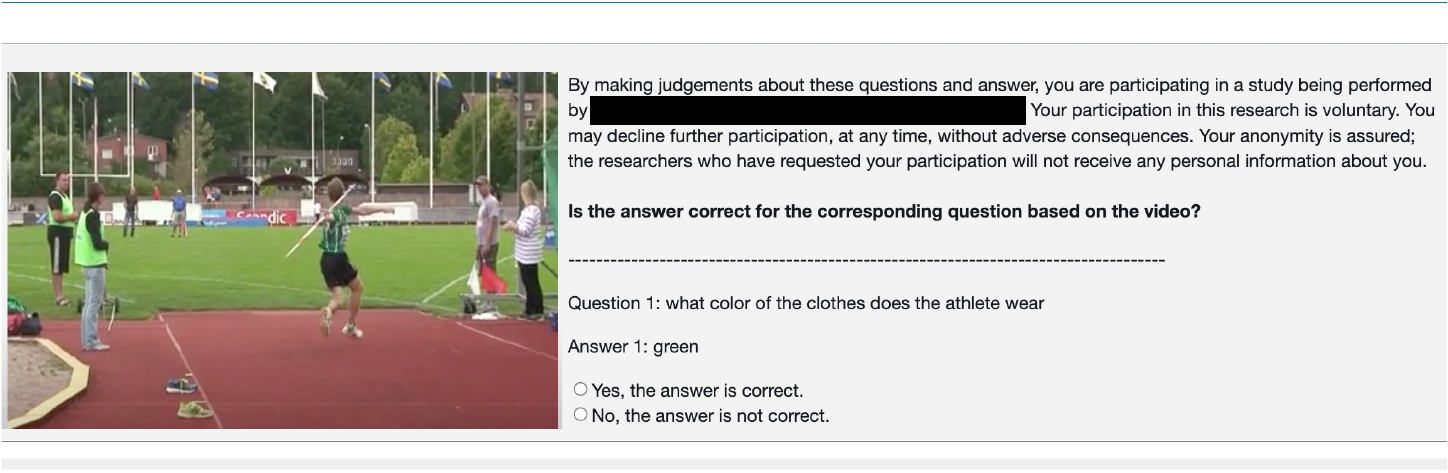}
\end{center}
\vspace{-5pt}
\caption{\small \textbf{Screenshot of Amazon Mechanical Turk we used for the video question answering experiment.} Workers are shown a video, three questions, and the answer to each question. The answers are generated by different methods. The workers are not told which method generates each answer. The workers are asked to select ``yes'' or ``no'' based on their measurement of whether the answer is correct for the given video and question. 
% If the answer matches the given video and question, the worker should choose ``yes''; otherwise, ``no''.
}
\label{apxfig:mturk}
\vspace{-5pt}
\end{figure}

To evaluate approaches on solving the zero-shot video question answering tasks, we ask workers from Amazon Mechanical Turk to evaluate the generated answer based on the video and the asked question. 
Before showing the questions and answers to the workers, we used the profanity check tool (\href{https://github.com/vzhou842/profanity-check}{\texttt{https://github.com/vzhou842/profanity-check}}) to remove the improper questions and answers.
As shown in \cref{apxfig:irb}, this experiment was approved by the Committee on the Use of Humans as Experimental Subjects.
A screenshot of the task is shown in \cref{apxfig:mturk}. The instructions shown to participants are listed as follows:

\begin{adjustwidth}{1.3cm}{}
    \textbf{Instructions:} By making judgments about these questions and answers, you are participating in a study being performed by [XXX]. Your participation in this research is voluntary. You may decline further participation, at any time, without adverse consequences. Your anonymity is assured; the researchers who have requested your participation will not receive any personal information about you.
\end{adjustwidth}

Given a video, a question, and a generated answer, the workers from Amazon Mechanical Turk measure whether the answer is correct for the given question and video. Each video shows three question-answer pairs (only one question-answer pair is shown in the screenshot). The answers are generated by different methods. The workers are not told which method generates each answer.
The workers are asked to choose ``yes'' or ``no''.
If the worker thinks the answer matches the given video and question, they should choose ``yes''; otherwise, ``no''.

% For fair comparisons, all the approaches answer the same 300 video questions, and each answer is evaluated by three different workers.
To control the quality, each task is evaluated by three different workers. The workers are required to have an approval rate greater than $98\%$.
Our test shows that each task takes around 10 seconds, but the workers are given up to one hour to complete each task. The workers are paid $\$0.05$ for finishing each task with an estimated hourly payment of $\$18$, more than the United States federal minimum wage. There are 33 workers in total who joined our experiment.

\end{document}